\documentclass[10pt,journal,compsoc]{IEEEtran}
\usepackage{amsmath,amsfonts}
\usepackage{array}
\usepackage[nocompress]{cite}
\usepackage[numbers,sort,square]{natbib}
\usepackage{url}
\usepackage{graphicx}
\def\BibTeX{{\rm B\kern-.05em{\sc i\kern-.025em b}\kern-.08em
    T\kern-.1667em\lower.7ex\hbox{E}\kern-.125emX}}
\usepackage{balance}

\usepackage[T1]{fontenc}
\usepackage[utf8]{inputenc}
\usepackage{microtype}
\usepackage{fontawesome5}
\usepackage{tabularx}
\usepackage{nicefrac}
\usepackage{float}
\usepackage[export]{adjustbox}
\usepackage[most]{tcolorbox}
\usepackage{array}
\usepackage{soul}
\usepackage[super]{nth}
\usepackage{mdframed}
\usepackage{inconsolata}
\usepackage{bm}
\usepackage{mathrsfs}
\usepackage{epsfig}
\usepackage{graphicx}
\usepackage{booktabs}
\usepackage{verbatim}
\usepackage{tipa}
\usepackage{algorithm}
\usepackage{url}
\usepackage{amsfonts}       
\usepackage{nicefrac}       
\usepackage{xcolor}         
\usepackage{tcolorbox}
\usepackage{algorithm}
\usepackage{listings}
\usepackage{inconsolata}
\usepackage{bm}
\usepackage{amsthm,amsmath,amssymb}
\usepackage{mathrsfs}
\usepackage{epsfig}
\usepackage{booktabs}
\usepackage{verbatim}
\usepackage{enumitem,kantlipsum}
\usepackage{lipsum}
\usepackage{tipa}
\usepackage{algorithmicx}
\usepackage{algpseudocode}
\usepackage{multirow,makecell}
\usepackage{arydshln}
\usepackage{colortbl}

\usepackage{color} 
\usepackage{dsfont}
\usepackage{tikz}
\usepackage{wrapfig}
\usepackage{xspace}
\usepackage{ifthen}
\usepackage{subcaption}
\usepackage{pdflscape}
\usepackage{pifont}
\usepackage{bbm}
\usepackage{dsfont}
\usepackage{graphicx}
\usepackage{subcaption}  
\usepackage{caption}

\newtcolorbox{AIbox}[2][]{aibox,title=#2,#1}

\usepackage{hyperref}
\tcbset{
  aibox/.style={
    width=450pt,
    top=10pt,
    colframe=black,
    colbacktitle=black,
    enhanced,
    center,
    attach boxed title to top left={yshift=-0.1in,xshift=0.15in},
    boxed title style={boxrule=0pt,colframe=white,},
  }
}
\definecolor{forestgreen}{rgb}{0.13, 0.55, 0.13}

\usepackage{color} 
\definecolor{mypink2}{RGB}{219, 48, 122}
\definecolor{orange}{RGB}{255, 147, 00}
\definecolor{jrcolor}{RGB}{100, 150, 225}
\definecolor{jrcomment}{RGB}{70, 200, 150}
\definecolor{grey}{RGB}{166, 166, 166}

\definecolor{mygreen}{HTML}{3cb44b}

\DeclareMathOperator*{\argmin}{arg\,min}

\definecolor{mypink2}{RGB}{219, 48, 122}
\definecolor{orange}{RGB}{255, 147, 00}
\definecolor{jrcolor}{RGB}{100, 150, 225}
\definecolor{jrcomment}{RGB}{70, 200, 150}

\NewDocumentCommand{\myparagraph}{m}{%
  \par\vspace{1.5ex}%
  \noindent\textbf{#1}\hspace{0.5em}%
}

\usepackage{cleveref}
\crefname{section}{§}{§§}
\Crefname{section}{§}{§§}

\definecolor{NavyBlue}{rgb}{0.1, 0.4, 0.8}
\hypersetup{colorlinks = true, linkcolor = NavyBlue,
            urlcolor  = gray,
            citecolor = NavyBlue,
            anchorcolor = NavyBlue}

\begin{document}

\title{SeqPE: Transformer with Sequential Position Encoding}

\author{Huayang Li, Yahui Liu, Hongyu Sun, Deng Cai, Leyang Cui, Wei Bi, \\ Peilin Zhao, Taro Watanabe
\thanks{Huayang Li, Hongyu Sun and Taro Watanabe are with Nara Institute of Science and Technology (NAIST), Nara, Japan. Yahui Liu is with Kuaishou Technology, Beijing, China. Deng Cai, Leyang Cui and Wei Bi are independent researchers in China. Peilin Zhao is with Tencent, Shenzhen, China. Taro Watanabe and Yahui Liu are co-corresponding authors.}
\thanks{E-mail: \{li.huayang.lh6, taro\}@is.naist.jp, yahui.cvrs@gmail.com}
}

\markboth{Preprint}%
{Shell \MakeLowercase{\textit{et al.}}: A Sample Article Using IEEEtran.cls for IEEE Journals}


\IEEEtitleabstractindextext{
\begin{abstract}
Since self-attention layers in Transformers are permutation invariant by design, positional encodings must be explicitly incorporated to enable spatial understanding. 
However, fixed-size lookup tables used in traditional learnable position embeddings (PEs) limit extrapolation capabilities beyond pre-trained sequence lengths. Expert-designed methods such as \textsc{ALiBi} and \textsc{RoPE}, mitigate this limitation but demand extensive modifications for adapting to new modalities, underscoring fundamental challenges in adaptability and scalability.
In this work, we present \textsc{SeqPE}, a unified and fully learnable position encoding framework that represents each $n$‑dimensional position index as a symbolic sequence and employs a lightweight sequential  position encoder to learn their embeddings in an end-to-end manner.
To regularize \textsc{SeqPE}'s embedding space, we introduce two complementary objectives: a contrastive objective that aligns embedding distances with a predefined position‑distance function, and a knowledge distillation loss that anchors out-of-distribution position embeddings to in-distribution teacher representations, further enhancing extrapolation performance. 
Experiments across language modeling, long‑context question answering, and 2D image classification demonstrate that SeqPE not only surpasses strong baselines in perplexity, exact match (EM), and accuracy--particularly under context length extrapolation--but also enables seamless generalization to multi-dimensional inputs without requiring manual architectural redesign. 
We release our code, data, and checkpoints at \url{https://github.com/ghrua/seqpe}.
\end{abstract}

\begin{IEEEkeywords}
Transformers, Position Encoding, Position Embeddings, Contrastive Learning, Multi-modality
\end{IEEEkeywords}
}

\maketitle
\IEEEdisplaynontitleabstractindextext
\IEEEpeerreviewmaketitle

\section{Instruction}
\label{sec:instruction}

\begin{figure*}[!t]
    \begin{center}
        \includegraphics[width=0.82\textwidth]{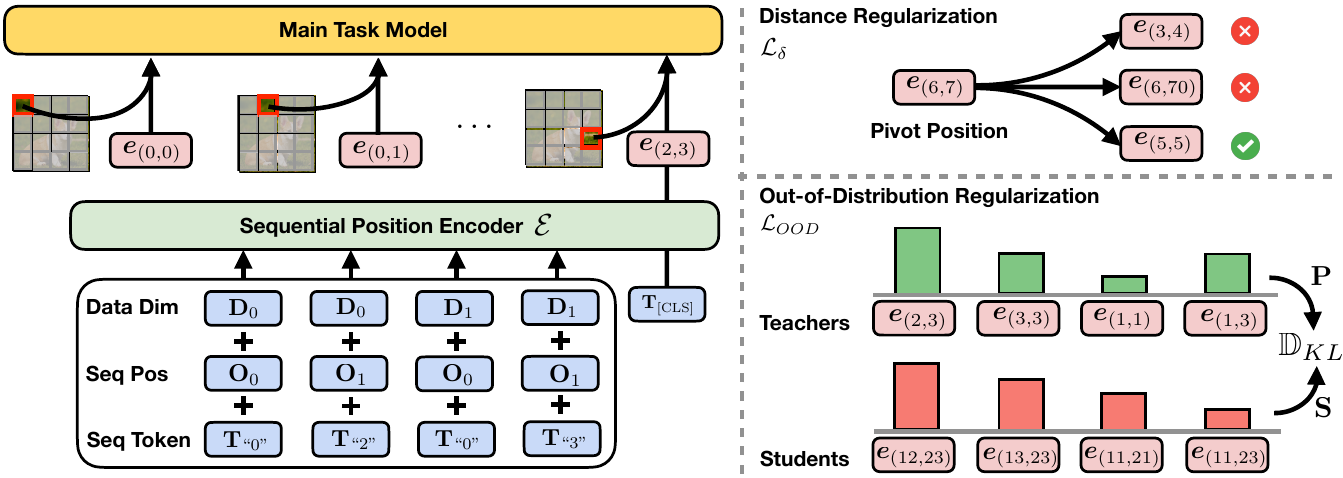}
    \end{center}
    \caption{Overall architecture and objectives for \textsc{SeqPE}. We use the representation for the (2, 3) position in an image with 4$\times$4 patches as an example, where SeqPE converts the (2, 3) position to a left-padded sequence (``0'', ``2'', ``0'', ``3'') and learn its representation by a sequential position encoder. The main purpose of the left-side padded ``0'' for each dimension is to ensure that digits with the same place value (\emph{e.g.}, units, tens) always occupy the same position in the sequence, resulting in a more consistent representation and improved training efficiency. With the sequential interface of positions, \textsc{SeqPE} can be easily extended to positions of longer context and varied dimensions. Moreover, two regularization objectives, agnostic to data dimensionality, are introduced to regularize the position representation space of \textsc{SeqPE}.\label{fig:illustration}}
\end{figure*}

\IEEEPARstart{T}he Transformer model~\cite{vaswani2017attention} has become the dominant framework for AI tasks across data with varied modalities, including text~\cite{radford2019language,brown2020language}, image~\cite{DBLP:conf/iclr/DosovitskiyB0WZ21,DBLP:conf/icml/TouvronCDMSJ21}, and video~\cite{girdhar2023imagebind}.
However, the self-attention mechanism in Transformers is inherently permutation-invariant, lacking an intrinsic understanding of input token order, an aspect crucial for capturing the semantics of sequential data. 
To address this, position encoding is introduced to inject order-sensitive information into the models~\cite{gehring2017convolutional,vaswani2017attention}, enabling Transformers to distinguish tokens based not only on their content but also on their position within the context.

Recently, there has been extensive research exploration on position encoding, which can be broadly categorized into two main directions. In the first direction, fixed or learnable position embeddings (PEs) are provided through the use of fixed-size lookup tables~\cite{radford2019language,devlin2019bert,raffel2020exploring,DBLP:conf/iclr/DosovitskiyB0WZ21}. 
Such PEs are typically added to the input text or vision embeddings as the input to transformer models. 
However, due to training constraints imposed by fixed maximum sequence lengths, such PEs usually exhibit poor extrapolation capacity to sequence lengths that exceed the length encountered during training.

In the second direction, positional information is injected by modifying attention scores with expert-designed weights or mapping functions~\cite{press2021train,almazrouei2023falcon,su2024roformer,touvron2023llama,yang2025qwen3,su2024roformer,heo2024rotary}. 
A key advantage of such methods is their ability to establish consistent, predetermined positional relationship patterns, enabling models to extrapolate to previously unseen sequence positions. 
However, they also come with two crucial limitations: (1) They primarily model ``distance decay'' or ``local emphasis'' but may fail to capture more subtle, non-monotonic or directional positional relationships that learnable embeddings could represent. (2) They use fixed, non-learned functional form for the bias, and lack adaptability for different datasets, tasks~\cite{li2024aria,zhou2024transfusion} or even different layers within the same model~\cite{haviv2022transformer}.

In this work, we propose an alternative scheme for learnable position encoding, called \textsc{SeqPE}, which employs a sequential position encoder to directly map a sequential representation--derived from an $n$-dimensional position--into a hidden vector representing that position. Concretely, we represent the $n$-dimensional position using three types of embeddings: sequence tokens, sequence positions, and data dimensions. The embedded sequence is then encoded using a lightweight encoder $\mathcal{E}$. An example of the position representation for 2D data is illustrated on the left-hand side of Figure~\ref{fig:illustration}.  By adopting this sequential representation as the new interface, our learnable position encoding method can easily encode unseen positions, thereby overcoming the limitation of standard learnable position encoding methods, which can only handle a fixed number of positions. 
Additionally, compared to PEs with expert-designed weights or mapping functions, our method offers superior adaptability to diverse tasks and enhanced extensibility to data across varying dimensionality.

To guarantee the stability of performance,  there is a critical challenge needs to be addressed: 
\textit{sequential models often struggle to distinguish semantic differences between lexically similar sequences}~\cite{bandel-etal-2022-lexical, jia-liang-2017-adversarial}. 
For example, while one might intuitively expect that closer positions would yield more similar representations, we find that, in the case of one-dimensional data, the embedding of ``100'' learned by \textsc{SeqPE} is closer to that of ``1000'' than to ``123'', because ``100'' and ``1000'' share higher lexical overlaps. To tackle this challenge, we introduce a contrastive-learning-based objective to regularize the representation space of encoded position embeddings. 
Specifically, given a predefined distance function $\delta$ between positions, we train \textsc{SeqPE} to ensure that pairs of positions $p$ and $p'$ that are closer under $\delta$ produce more similar embeddings, where $p$ and $p'$ may have arbitrary dimensionality.

In addition, although
the contrastive-learning strategy 
has the potential to generalize to unseen positions, its performance on the main task may degrade when evaluated on out-of-distribution (OOD) positions. 
This is because the embeddings of such OOD positions may shift into regions of the state space that were under-optimized for the main task objective. 
To mitigate this issue, under the assumption that the main task model is largely optimized on positions within the training context of $n$-dimensional data, we introduce a knowledge distillation objective. This objective leverages embeddings from the training context to guide the learning of embeddings for $n$-dimensional OOD positions.

We conducted extensive experiments to validate the effectiveness of \textsc{SeqPE} in three key aspects:
learnability and adaptability on various tasks, context extrapolation capability, and extensibility to higher-dimensional data. 
We evaluate our method in three tasks (\textit{i.e.}, language modeling, question answering, and image classification) under the context extrapolation setting.
We first evaluate on the language modeling task using the GPT-2 model~\cite{radford2019language}. Our \textsc{SeqPE} outperforms other baselines by at least 0.6 perplexity points on the Wikitext-103 test set. 
To assess adaptability, we fine-tuned the pre-trained language models on a long-context question answering task \cite{hsieh2024ruler}. Results show that, compared to other baselines, our method achieves at least 24.2 and 2.6 points of average performance improvement in perplexity and exact match (EM) metrics on the answer span, respectively. 
Finally, we extend our method to 2D image data and demonstrate that the ViT model \cite{DBLP:conf/icml/TouvronCDMSJ21, DBLP:conf/iclr/DosovitskiyB0WZ21} trained on ImageNet-1K with \textsc{SeqPE} surpasses other baselines by a minimum of 2.1 percentage points in classification accuracy.

In summary, our contributions are three-fold:
\begin{itemize}
\item We propose a novel scheme for learnable position encoding, called \textsc{SeqPE}, which replaces conventional index-based lookup schemes with sequential position encoder for position representation.
\item We introduce two training objectives to regularize the representation space of \textsc{SeqPE}: a distance alignment objective that matches embedding distances and positional distances, and an objective for generalization of out-of-distribution positions. Both objectives are agnostic to the dimensionality of data.
\item We perform comprehensive experiments across language modeling, long-context question answering, and 2D image classification tasks, demonstrating enhanced adaptability, superior context extrapolation capabilities, and effective extensibility to higher-dimensional data.
\end{itemize}

\section{Preliminary: Position Encoding in Transformers}
\label{sec:background}

Given an input data, \textit{e.g.}, a sequence of sub-words~\cite{devlin2019bert, radford2019language} or image patches~\cite{DBLP:conf/iclr/DosovitskiyB0WZ21}, the Transformer model~\cite{vaswani2017attention} first embeds it into a sequence of content embeddings $\mathbf{X} = (\boldsymbol{x}_1, \boldsymbol{x}_2, \dots, \boldsymbol{x}_L)\in\mathbb{R}^{L\times d}$, which has no position information yet. Then the Transformer model processes the information in $\mathbf{X}$ using multiple layers of self-attention-based neural networks. The core of the attention mechanism\footnote{For the simplicity of notions, we omit the multi-head part of the $\mathrm{Attention}$ function \cite{vaswani2017attention}.} for the $i$-th position in each layer is as follows:
\begin{align}
\mathbf{Q}, \mathbf{K}, \mathbf{V} &= \mathbf{H}\mathbf{W}^{\{q, k, v\}}, \\
\mathbf{A}_{i,j} & = \boldsymbol{q}_i^\top\boldsymbol{k}_j\\
\mathrm{Attention}(\mathbf{A}, \mathbf{V})_i &= \frac{\sum_{j=1}^L \exp( \mathbf{A}_{i,j} / \sqrt{d})\boldsymbol{v}_j}{\sum_{j=1}^L \exp( \mathbf{A}_{i,j} / \sqrt{d})},\label{eq:attn}
\end{align}
where $\mathbf{W}^{\{q,k,v\}} \in \mathbb{R}^{d\times3d}$ is the parameter for the linear transformation, $\mathbf{H} \in \mathbb{R}^{L\times d}$ is the input of this layer, and $\mathbf{A}\in\mathbb{R}^{L\times L}$ contains the non-normalized attention scores between any pair of positions. Notably,  $\mathbf{H} = \mathbf{X}$ for the first layer input if no position information is used (\textit{e.g.}, \textsc{NOPE} method \cite{kazemnejad2023impact, yang2025RoPE}).
%
To equip the model with a sense of order at the input level, the Absolute Position Embedding (APE) method \cite{gehring2017convolutional, vaswani2017attention} is proposed by adding PEs to the input content embeddings:
\begin{equation}
    \mathbf{X}' = (\boldsymbol{x}_1 + \boldsymbol{e}_1, \boldsymbol{x}_2+\boldsymbol{e}_2, \dots, \boldsymbol{x}_L+\boldsymbol{e}_L),
\end{equation}
where $\boldsymbol{e}_i$ is the position embedding for the $i$-th position, which could be sinusoidal-based fixed weights \citep{vaswani2017attention} or learnable embeddings \citep{radford2019language, devlin2019bert}. Afterwards, $\mathbf{H} = \mathbf{X}'$ is used as the first layer input for the APE method \cite{gehring2017convolutional}.

Differently, the \textsc{ALiBi} method \citep{press2021train} does not add the position information into $\mathbf{X}$ (\textit{i.e.}, $\mathbf{H} = \mathbf{X}$ for the first layer input of the Transformer model), but injects the position information into the attention mechanism:
\begin{align}
\mathbf{A}_{i,j}^\text{ALiBi} &= \boldsymbol{q}_i^\top\boldsymbol{k}_j + \boldsymbol{M}_{i,j} 
\end{align}
where $\mathbf{A}_{i,j}^\text{ALiBi}$ will be used for the attention function in Equation~\ref{eq:attn}. The $\mathbf{M}_{i, j}$ is the expert-designed bias for text data:
\begin{equation}
\mathbf{M}_{i,j} = 
\begin{cases}
- m \cdot (i - j), & \text{if } i \geq j, \\
- \infty, & \text{otherwise}, \label{eq:alibi_matrix}
\end{cases}
\end{equation}
where $m$ is a head-specific scale value.

Alternatively, the core of \textsc{RoPE} \cite{su2024roformer} is the rotary matrix $\mathbf{R}_{\Theta}$, which is designed based on Euler's formula, where $\Theta$ represents a pre-defined frequencies vector. The $\mathbf{R}_{\Theta}$ is directly applied to the query $\boldsymbol{q}_i$ and key $\boldsymbol{k}_j$ as follows:
\begin{align}
\mathbf{A}_{i,j}^\text{RoPE} &= \boldsymbol{q}_i^\top\mathbf{R}_{\Theta,i}^\top\mathbf{R}_{\Theta,j}\boldsymbol{k}_j
\end{align}
where a property of the rotary matrix is that $\mathbf{R}_{\Theta, j-i}=\mathbf{R}_{\Theta, i}^\top\mathbf{R}_{\Theta, j}$, \textit{i.e.}, modeling the relative relationship between positions.

\section{Methodology}
\label{sec:methods}

In this section, we introduce our general position encoding method that involves three key properties within a single model: the learnability and adaptability across different tasks; context extrapolation (\textit{i.e.}, generalization capability to sequences longer than those encountered during training); and extensibility to data of arbitrary dimensions with minimal additional manual efforts.

\subsection{Overview} To achieve these goals, we propose an alternative framework for the position encoding, which employs a sequential position encoder (\textsc{SeqPE}) that directly maps an $n$-dimensional position into a hidden vector representing that position via a sequential modeling process.  As illustrated in Fig.~\ref{fig:illustration}, when representing the position $(2, 3)$ in an image with $4 \times 4$ patches, \textsc{SeqPE} first converts it into a left-padded sequence (``0'', ``2'', ``0'', ``3'') and then uses a multi-layer sequential encoder~\cite{vaswani2017attention} to transform the sequence into the hidden representation $\boldsymbol{e}_{(2,3)}$ of the position. Details of sequential representation and position encoder are provided in Section~\ref{sec:pos_encoder}. 

This sequential modeling approach offers several advantages. First, \textsc{SeqPE} can leverage fixed learning parameters to model a dynamic number of input positions. In contrast, standard learnable position embeddings typically rely on a lookup table whose size grows linearly with the training context length, consuming extensive memory when training on long-context data. Moreover, the size of the lookup table is fixed, making it difficult to extend to unseen positions in longer contexts. Second, the new input interface in \textsc{SeqPE}, a sequence of position tokens, is unified for data of any dimensionality, including hybrid-dimensional data~\cite{li2024aria,zhou2024transfusion} such as interleaved text and images. As a result, \textsc{SeqPE} can be readily extended to data of varying dimensions without the need for manual adjustments, unlike many expert-crafted position encoding methods \cite{su2024roformer, press2021train}\footnote{For instance, extending the \textsc{RoPE} method from 1D to 2D requires careful expert design~\cite{heo2024rotary}.}.

However, simply integrating the position embeddings learned by \textsc{SeqPE} into the main task model, \textit{e.g.}, for language modeling or image classification, does not always yield satisfactory results in our preliminary experiments. One limitation is that the representation space learned by \textsc{SeqPE} struggles to distinguish distance differences between lexically similar position sequences. For example, the embedding of ``100'' learned by non-regularized \textsc{SeqPE} is closer to that of ``1000'' than to ``123'', which introduces non-beneficial bias to the main task model. Additionally, although \textsc{SeqPE} has the potential to generalize to unseen positions, its performance on the main task may degrade when evaluated on out-of-distribution (OOD) positions, because embeddings of such OOD positions may shift into regions of the state space that were under-optimized for the main task objective.
In the rest of this section, we introduce two position objectives to regularize the representation space of \textsc{SeqPE}, solving the two aforementioned challenges: embedding distances (Section~\ref{sec:pos_semantics}) and out-of-distribution generalization (Section~\ref{sec:pos_transferbility}). It is worth noting that all the proposed regularization objectives are agnostic to the data dimensionality, although we frequently use the one-dimensional case for explanation.

\subsection{Sequential Position Encoding}
\label{sec:pos_encoder}

Our proposed \textsc{SeqPE} aims to 
maps an $n$ dimensional position into a $d$ dimensional hidden vector using a sequential model parameterized by $\theta$: $f_\theta\colon \mathbb{Z}^n \to \mathbb{R}^d$. As show in Fig.~\ref{fig:illustration}, the encoding process $f_\theta$ includes: 1) representing the position $p$ using a sequence of hidden vectors; 2) encoding the sequence into $\boldsymbol{e}_p$  using a sequential model.

\myparagraph{Position Representation}
Given an $n$ dimensional position $p$, \textit{e.g.}, $(2, 3)$-th patch of a 2D image, we first convert it to a concatenated left-padded sequence $\boldsymbol{s} = (s^0_0, \dots, s^0_{k-1}, \dots, s^{n-1}_0, \dots s^{n-1}_{k-1})$, \textit{e.g.}, $\boldsymbol{s} =$(``0'', ``2'', ``0'', ``3''), where $k$ is the maximum number of digits for each dimension. The sub-sequence for each dimension with less than $k$ digits will be padded with ``0'' from the left-hand side to ensure that digits with the same place value (\emph{e.g.}, units, tens) always occupy the same position in the sequence, resulting in a more consistent representation and improved training efficiency. For each token $s^i_j$ in the sequence, we represent it using the addition of three embeddings:
$$
\mathbf{U}^i_j = \mathbf{T}_{s^i_j} + \mathbf{O}_{j} + \mathbf{D}_i,
$$
where $\mathbf{T}\in\mathcal{R}^{(b+1)\times d}$ is the embedding matrix for all the possible sequence tokens from positions\footnote{The value of $b$ is determined by the base of the number system we choose; for example, $b$ is 10 in the decimal system and 16 in the hexadecimal system. We use the decimal system by default.} and the special token ``[CLS]'', $\mathbf{O}\in\mathcal{R}^{k\times d}$ is the embedding for each token's position in the sequence, and $\mathbf{D}\in\mathcal{R}^{n\times d}$ is the embedding for data dimensions. Notably, our system can represent $(b^k)^n$
positions using $b+n+k$ embeddings. Thus, the number of embeddings required by our system grows logarithmically with the number of positions, whereas it grows linearly in lookup-table-based methods~\cite{radford2019language, devlin2019bert}. The $\mathbf{U}\in \mathcal{R}^{nk\times d}$ is the hidden representation of the sequence $\boldsymbol{s}$. As illustrated in Fig.~\ref{fig:illustration}, we set $n=2$, $k=2$, and $b=10$ to represent the positions of patches in 2D images from $(0, 0)$ to $(99, 99)$. The use case for 1-dimensional data is shown in Appendix \ref{app:seqpe_embed_1d}.

\begin{figure}[t]
    \centering
    \begin{subfigure}[b]{0.45\columnwidth}
        \includegraphics[width=1.0\columnwidth]{"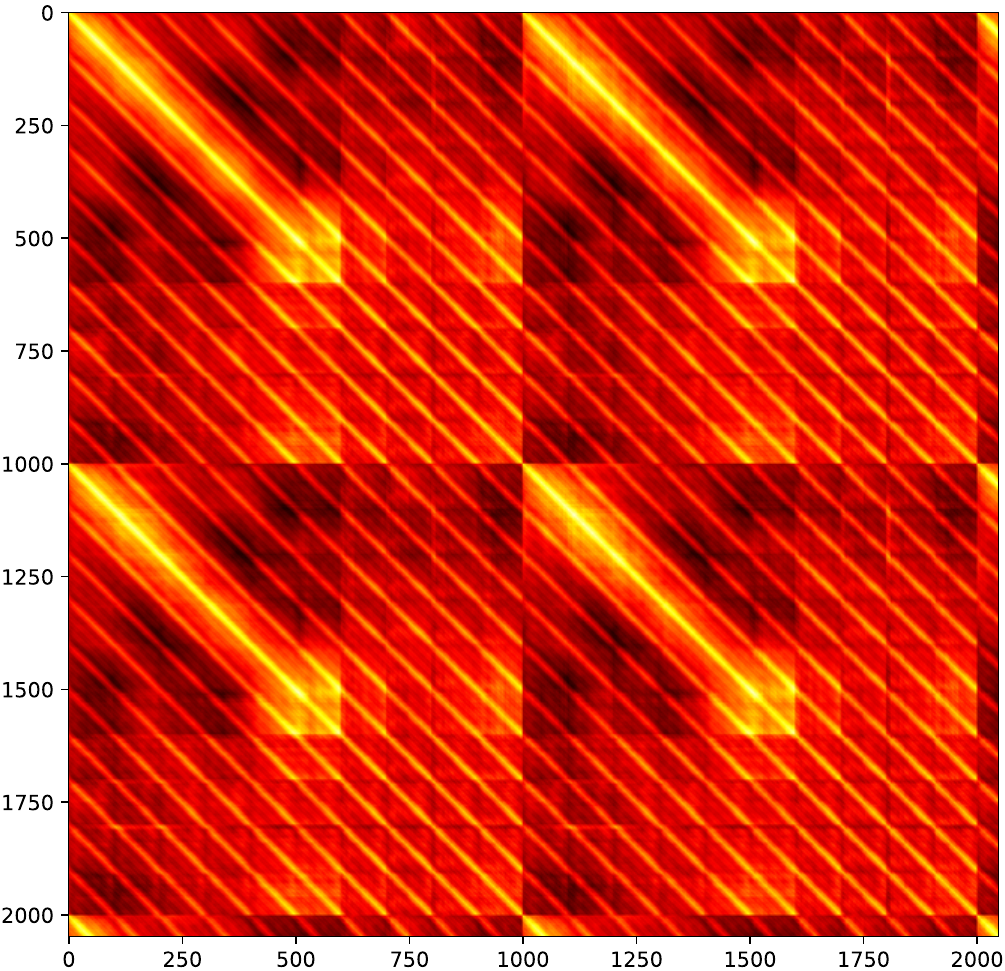"}
        \caption{\textsc{SeqPE} w/o $\mathcal{L}_\delta$}
        \label{fig:pos_semantics_wo_ct_loss}
    \end{subfigure}
    \hspace{0.05\columnwidth}
    \begin{subfigure}[b]{0.45\columnwidth}
        \includegraphics[width=1.0\columnwidth]{"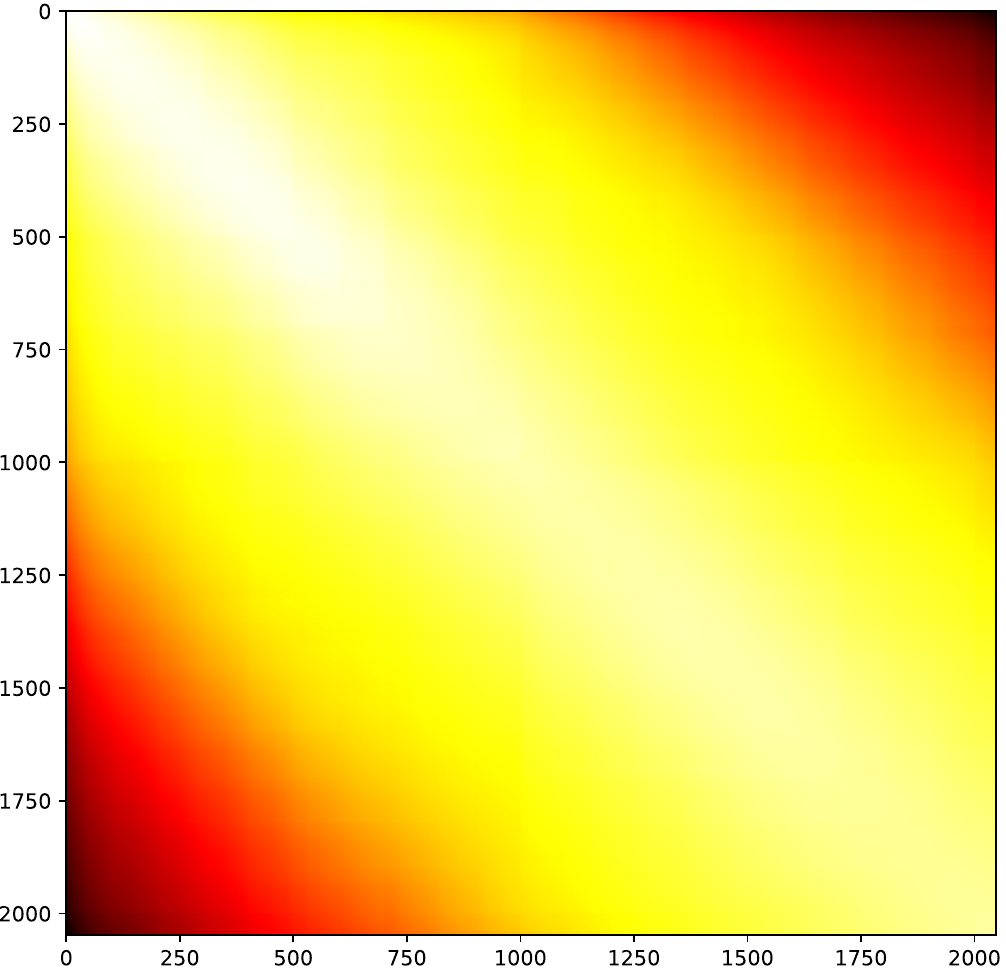"}
        \caption{\textsc{SeqPE} w/ $\mathcal{L}_\delta$}
        \label{fig:pos_semantics_w_ct_loss}
    \end{subfigure}
    \caption{Heatmap for the values of dot product between position embeddings from 0 to 2048 for text data. Brighter area indicate higher similarity between two position embeddings. The regularization objective $\mathcal{L}_\delta$ makes \textsc{SeqPE} learn smoother representations. Implementation details are in Appendix \ref{app:visual_impl}.}
    \label{fig:pos_semantics}
\end{figure}

\myparagraph{Encoder Architecture}
To effectively represent positional information, we introduce a sequential modelling network. Any sequential modeling network can be used for this task. In our work, we use a lightweight Transformer model \cite{vaswani2017attention} with $N$ layers (\emph{e.g.,} $N=2$) and causal attention\footnote{In our preliminary study, the bi-directional and causal attention (\textit{i.e.}, decoder-only) achieves almost the same performance}. We add an additional ``[CLS]'' token at the end of each position sequence, and use its encoder output as the final representation for the position $p$:
\begin{equation}
    \boldsymbol{e}_p = \mathcal{E}(\mathbf{U} \oplus \mathbf{T}_{\mathrm{[CLS]}}; \pmb{\theta}),\label{eq:enc}
\end{equation}
where $\mathcal{E}(\cdot ; \pmb{\theta})$ refers to our proposed sequential position encoder with trainable parameters $\pmb{\theta}$, and $\oplus$ is the concatenation operation between vectors.

\begin{table*}[!htp]
\centering 
\caption{Perplexity ($\downarrow$) on the test set of Wikitext-103 with varied context lengths. We train a language model from scratch on sequences with $L=512$ tokens. The best and second best results are represented by \textbf{bold text} and \underline{underline}, respectively.\label{tab:wt103}}
\resizebox{0.7\textwidth}{!}{
\begin{tabular}{lrrrrrrr}\toprule
\textbf{Model}  & \textbf{512}          & \textbf{1K}   & \textbf{2K}    & \textbf{4K}   & \textbf{8K} & \textbf{16K}   & \textbf{AVG.}           \\\midrule
\textsc{NOPE}       &    20.40 & 34.98 & 116.11 &	289.73 & 536.29 & 881.56 &        313.17    \\
\textsc{APE-Sin}      &    23.37 & 77.86 & 628.68 & 2.3e3 & 5.0e3 & 9.4e3 &  2.9e3                                  \\
\textsc{APE-Learn}      &   20.30 & 21.36 & 37.70 & 86.90 &	152.53 &	230.63 &   91.57                       \\\hdashline
\textsc{RoPE}-$10^4$       &  19.44 & 35.25 & 	113.17 &	252.15 &	477.89 &	745.52 &	273.90         \\
\textsc{RoPE}-$10^6$       &  19.46  &   30.18 &	92.01 & 	201.57 &	332.04 &	539.20 &   202.41                  \\
\textsc{RoPE}-$10^8$       &  \textbf{19.42} &	42.92 & 144.71 &	341.40 &	541.17 & 786.08 &     371.25                \\\hdashline
\textsc{ALiBi}       &   20.06 & 	\underline{19.06} & \underline{18.65} &	\underline{18.72} &\underline{19.36} &	\underline{21.39} & \underline{19.54}               \\\midrule
\textsc{SeqPE}       &    \underline{19.65} &	\textbf{18.74} &	\textbf{18.51} & 	\textbf{18.58} & \textbf{18.86} & \textbf{19.37} & \textbf{18.95} \\\bottomrule  
\end{tabular}
}
\end{table*}

\subsection{Regularization for Embedding Distances}
\label{sec:pos_semantics}

Previous research reveals that sequential models often struggle to distinguish semantic differences between lexically similar sequences \cite{bandel-etal-2022-lexical, jia-liang-2017-adversarial}. Similarly, the representation space of the position encoder (\textit{i.e.}, Eq.~(\ref{eq:enc})) may become biased toward positions with similar input sequences. To evaluate this limitation of sequential modeling, we visualize the dot product between position embeddings (from 0 to 2048) generated by a non-regularized \textsc{SeqPE}, trained jointly with a \textsc{GPT-2} language model~\cite{radford2019language}. As shown in Fig.~\ref{fig:pos_semantics}(a), in the case of 1-dimensional text data, positions separated by intervals of 100 or 1000 are often closer in the representation space than adjacent positions. This unnatural bias can negatively impact both training and extrapolation performance. Further implementation details for visualization are provided in Appendix \ref{app:visual_impl}.

To address this issue, we propose an objective to regularize the embedding distances of positions. Specifically, for a given distance function on $n$-dimensional positions $\delta(p, p')\colon \mathbb{Z}^n\times\mathbb{Z}^n \to \mathbb{R}_{\ge 0}$, we assume that any positions $p$ and $p'$ that are closer under $\delta$ are expected to yield more similar embeddings under the encoding function $f_\theta$. Inspired by contrastive learning \cite{chen2020simple,gao-etal-2021-simcse}, our objective $\mathcal{L}_\delta$ is designed as follows:
\begin{equation}
    \mathcal{L}_\delta = -\log \frac{\exp(\boldsymbol{e}_{p} \cdot \boldsymbol{e}_{p^{+}})}{\sum_{p'\in \mathcal{C}}\exp(\boldsymbol{e}_{p}\cdot \boldsymbol{e}_{p'})\label{eq:contrastive}}
\end{equation}
where $\mathcal{C}_{\delta}$ is a set of randomly collected positions and $p^+ = \argmin_{p'} \{\delta(p, p') | p' \in \mathcal{C}_{\delta}\}$. We simply choose Euclidean distance for $\delta$ in this work. However, exploration for more distance functions, \textit{e.g.}, customized distance for interleaved data with varied dimensions, will be interesting for future works.  In Fig.~\ref{fig:pos_semantics}(b), we demonstrate that training a GPT-2 language model~\cite{radford2019language} with $\mathcal{L}_\delta$ get more reasonable representation space.

\subsection{Regularization for Out-of-Distribution Positions}
\label{sec:pos_transferbility}
We first define the in-distribution training context as 
$\mathcal{D}_{train} = \prod_{i=0}^{n-1} [0, L_i) \cap \mathbb{Z}^n,$
which is the Cartesian product of $n$ sets of position ranges, where $[0, L_i)$ denotes the in-distribution range along the $i$-th dimension. For example, the training positions for text data are drawn from $[0, L_0)$, while for 2D image data, they are drawn from $[0, L_0) \times [0, L_1)$, \emph{i.e.}, the set of positions $\{(x, y) \mid x \in [0, L_0),\ y \in [0, L_1)\}$.

Although our regularization for embedding distances (Section~\ref{sec:pos_semantics}) has the potential to generalize to unseen positions, the representation space for out-of-distribution (OOD) positions may be under-optimized for the main task objective. Specifically, the positions within the training context $\mathcal{D}_{train}$ can be optimized for both the main task (\emph{e.g.}, text perplexity) and the $\mathcal{L}_{\delta}$ objectives, but positions outside the training context $\mathcal{D}_{train}$ are generally only optimized with respect to $\mathcal{L}_{\delta}$, lacking the regularization for main task performance.

Here, we propose a knowledge-distillation-based objective $\mathcal{L}_{OOD}$ to gently bridge the gap between positions within and beyond the training context. Concretely, our idea is to use the representations of positions in training context $\mathcal{D}_{train}$ to teach the positions in a shifted region of OOD context $\mathcal{D}_{z} = \prod_{i=0}^{n-1} [z_i, L_i+z_i) \cap \mathbb{Z}^n$, where $|\mathcal{D}_{z}| = |\mathcal{D}_{train}|$, $z = (z_0, \dots, z_{n-1})$, and $z_i$ is a random shift for the start position at the $i$-th dimension. More concretely, we randomly sample a set of teacher positions $\mathcal{C}_{OOD} = \{p_0, p_1, \dots p_{m-1}\}$ without replacement, where $p_j \in \mathcal{D}_{train}$, and calculate similarity matrices $\mathbf{P}, \mathbf{S} \in \mathbb{R}^{m\times m}$ for teacher and student positions respectively:
\begin{equation}
\begin{aligned}
    \mathbf{P}_{i, j} &= \frac{\exp(\boldsymbol{e}_{p_i}\cdot\boldsymbol{e}_{p_j})}{\sum_{k=1}^m \exp(\boldsymbol{e}_{p_i}\cdot\boldsymbol{e}_{p_k})}, \nonumber\\
    \mathbf{S}_{i, j} &= \frac{\exp(\boldsymbol{e}_{p_i+z}\cdot\boldsymbol{e}_{p_j+z})}{\sum_{k=1}^m \exp(\boldsymbol{e}_{p_i+z}\cdot\boldsymbol{e}_{p_k+z})},\nonumber
\end{aligned}\label{eq:pq_matrices}
\end{equation}
where $p_j+z = (p_{j, 1}+z_1, \dots, p_{j, n-1}+z_{n-1})$. Thereafter, we optimize our position encoder by the knowledge distillation objective:
\begin{equation}
    \mathcal{L}_{OOD} = \mathbb{D}_{KL}(\mathrm{SG}(\mathbf{P}) || \mathbf{S}),\label{eq:distill}
\end{equation}
where $\mathbb{D}_{KL}$ is the Kullback–Leibler (KL) divergence and the $\mathrm{SG}(\cdot)$ is the stop-gradient operation \cite{van2017neural}. In addition, we find that training the main task model on shifted positions $\mathcal{D}_{z}$ rather than $\mathcal{D}_{train}$ for a small portion of data can further minimize the $\mathcal{L}_{OOD}$ and stabilize the extrapolation performance~\footnote{We do not train on larger context size, but only a random shift of start position. Moreover, many popular relative PE methods are invariant to the shift $z$~\cite{press2021train, su2024roformer, heo2024rotary}, as explained in Appendix \ref{app:random_shift}, ensuring a fair comparison.}. 

\myparagraph{Discussion} We expect the $\mathcal{L}_{OOD}$ objective to transfer the relative patterns of in-distribution PEs to out-of-distribution PEs, thereby enhancing extrapolation performance on longer contexts. However, a trivial solution exists that minimizes $\mathcal{L}_{OOD}$ by collapsing the embeddings, \emph{i.e.}, learning position embeddings $\boldsymbol{e}_{p}$ to be identical to their shifted counterparts $\boldsymbol{e}_{p+z}$. This behavior undermines extrapolation, as it destroys meaningful relative positional differences. Fortunately, the embedding distance regularization term, $\mathcal{L}_\delta$ in Equation~\ref{eq:contrastive}, discourages such collapse. That is, for positions satisfying $\delta(p, p+z') < \delta(p, p+z)$, \emph{i.e.},  $p$ is closer to $p+z'$ than to $p+z$, the objective $\mathcal{L}_\delta$ encourages
$$\exp(\boldsymbol{e}_{p} \cdot \boldsymbol{e}_{p+z^\prime}) > \exp(\boldsymbol{e}_{p} \cdot \boldsymbol{e}_{p+z}).$$
As a result, making $\boldsymbol{e}_{p+z}$ identical to $\boldsymbol{e}_{p}$, which would imply $\exp(\boldsymbol{e}_{p} \cdot \boldsymbol{e}_{p+z^\prime}) \le \exp(\boldsymbol{e}_{p} \cdot \boldsymbol{e}_{p+z})$, is penalized. Moreover, our ablation study in Section~\ref{subsec:ablation} confirms that jointly optimizing $\mathcal{L}_{OOD}$ and $\mathcal{L}_\delta$ leads to better extrapolation performance than optimizing $\mathcal{L}_{OOD}$ alone.

\subsection{Training \& Inference}
\myparagraph{Training} Here we explain how to train the \textsc{SeqPE} along with the main task model. Most methods that integrate position embeddings (PE) into the main task model are designed specific for their own PEs such as rotary operation in \textsc{RoPE}~\cite{su2024roformer}. 
To verify the universality of our proposed scheme, we experimentally confirm that our method is not constrained by any specific integration approach.
We abstract three types of general integration methods, \textit{i.e.}, \textsc{AttnSum}, \textsc{AttnMul}, and \textsc{AttnBias}. For different tasks, we can flexibly select the most appropriate one based on the performance on the validation dataset.

Specifically, given $L$ consecutive positions (\textit{e.g.}, from $0$ to $L$), we first encode them into a embedding matrix $\mathbf{E} \in \mathbb{R}^{L\times d}$ using \textsc{SeqPE}, and then map $\mathbf{E}$ to query and key position embeddings, respectively:
\begin{equation}
\mathbf{E}^{q}, \mathbf{E}^{k} = \mathbf{E}\mathbf{W}^{\{q,k\}}_{PE}, \label{eq:pe_qk_map}
\end{equation}
where $\mathbf{W}^{\{q,k\}}_{PE} \in \mathbb{R}^{d\times2d}$ can be applied in each layer of the main task model, following the implementation in the Transformer model \cite{vaswani2017attention}. We leverage the $\mathbf{E}^{q}$ and $\mathbf{E}^{k}$ to implement three types of non-normalized attention scores between the $i$-th and $j$-th positions as follows:
\begin{align}
\mathbf{A}^\text{AttnSum}_{i,j} &= (\boldsymbol{q}_i+\boldsymbol{e}^q_i)^\top(\boldsymbol{k}_j+\boldsymbol{e}^k_j) \\
\mathbf{A}^\text{AttnMul}_{i,j} &= (\boldsymbol{q}_i\odot\boldsymbol{e}^q_i)^\top(\boldsymbol{k}_j\odot\boldsymbol{e}^k_j) \\
\mathbf{A}^\text{AttnBias}_{i,j} &= \boldsymbol{q}_i^\top\boldsymbol{k}_j +{\boldsymbol{e}^q_i}^\top\boldsymbol{e}^k_j
\end{align}
where $\odot$ is the Hadamard product. The un-normalized attention score matrices can be integrated into the $\mathrm{Attention}(\mathbf{A}, \mathbf{V})$ in Equation~\ref{eq:attn}. 
In addition, the ultimate objective for training with SeqPE is as follows:
\begin{equation}
    \mathcal{L} = \mathcal{L}_{main} + \alpha\mathcal{L}_\delta + \beta\mathcal{L}_{OOD}\label{eq:final_obj}
\end{equation}
where $\alpha$ and $\beta$ are two hyper-parameters, and $\mathcal{L}_{main}$ is the main task objective, \textit{e.g.}, cross entropy of the next-token prediction distribution~\cite{radford2019language} or image classification~\cite{DBLP:conf/iclr/DosovitskiyB0WZ21}.  

\myparagraph{Inference} Our \textsc{SeqPE} method is inference-friendly because the embeddings for identical positions can be pre-encoded by the encoder $\mathcal{E}$ only once. In other words, we can construct an efficient lookup table, as in standard methods, by pre-computing the embeddings for all positions. This not only reduces inference latency but also makes our method compatible with most existing inference engines for large language models \cite{kwon2023efficient, NEURIPS2024_724be447}.

\begin{table*}[]
\centering  
\caption{Perplexity and Exact-Match (EM) score on the validation set of long-context RULER-SQuAD dataset \cite{hsieh2024ruler}. We fine-tuned corresponding model in Section~\ref{sec:exp_lm} on the RULER-SQuAD training dataset with 1024 context length. The perplexity is calculated only based on the golden-truth answer span after the input prompt. The EM score is based on the tokens generated by greedy search. The best and second best results are represented by \textbf{bold text} and \underline{underline}, respectively. \label{tab:qa}}
\resizebox{0.8\textwidth}{!}{
\begin{tabular}{lrrrrr}\toprule
\textbf{Model}  &  \textbf{1K}   & \textbf{2K}    & \textbf{4K}   & \textbf{8K} &  \textbf{AVG.}           \\\hline\hline
\multicolumn{6}{c}{Perplexity$\downarrow$ / EM$\uparrow$} \\\hline\hline
\textsc{NOPE}  & 6.07 / 29.2 & 236.53 /\ \ \  3.6 & 967.89	/\ \ \  0.6 & 1.5e4 / 0.0 &  698.94 /\ \ \  8.3
		 \\
\textsc{APE-Sin} & 7.66 / 26.2 &	577.48 /\ \ \ 1.8 & 1.5e4 /\ \ \   0.0	& 1.5e6 / 0.0       & 4.3e4 /\ \ \  7.0                           \\
\textsc{APE-Learn}  & 6.79 / 25.0	& 15.66 / \underline{11.8} & 41.30 /\ \ \   6.2& 365.49 / 0.2   & 107.31 / 10.8                           \\
\textsc{RoPE}-$10^6$  	& 5.62 / \textbf{28.0} & 23.00 /\ \ \  9.0 & 	314.77 /\ \ \  3.0 & 963.48 /	0.8   & 326.72 / 10.2                  \\
\textsc{ALiBi}  & \textbf{5.49} /  23.8 & \underline{8.07} / 11.4 &	\underline{9.52} /\ \ \  \underline{8.0} &	\underline{121.14} / \underline{2.0}   &  \underline{36.06} / \underline{11.3}        \\\midrule
\textsc{SeqPE}  	&  \underline{5.56} / \underline{26.8} &	\textbf{7.64} / \textbf{15.0} & \textbf{7.58} / \textbf{11.4} &  \textbf{28.60} / \textbf{2.6}   & \textbf{12.34} / \textbf{13.9}  \\\bottomrule  
\end{tabular}
}
\end{table*}

\section{Experiments}
\label{sec:experiments}
We conducted extensive experiments to evaluate the three motivations of our \textsc{SeqPE}, \textit{i.e.}, learnability and adaptability, extrapolation, and extension to data with varied dimensions. 
Most \textsc{SeqPE} setups remain consistent across experiments. When representing positions, we use a decimal system (\textit{i.e.}, $b=10$). We represent positions for 1-dimensional text data using $k=5$ digits, and for 2-dimensional image data using $k=2$ digits per dimension. We use $N=2$ for the number of layers in architecture of the Transformer-based~\cite{vaswani2017attention} position encoder $\mathcal{E}$ with weight decay value setting to $0$. The weights $\alpha$ and $\beta$ in Equation~\ref{eq:final_obj} are both set to $0.1$ based on validation performance. Both the sizes of sampled positions for $\mathcal{C}_{\delta}$ and $\mathcal{C}_{OOD}$ are set to 32. We set the batch sizes for the two regularization objectives $\mathcal{L}_{\delta}$ and $\mathcal{L}_{OOD}$ to 32 (distinct from the main task). More implementation details are shown in Appendix~\ref{app:seqpe_impl}. Additionally, $10\%$ of the main task data is trained with shifted start positions to reduce the knowledge distillation loss in Equation~\ref{eq:distill}. This shift affects only the start position without changing context length, making it a no-op for strong baselines like \textsc{RoPE} \cite{su2024roformer, heo2024rotary} and \textsc{ALiBi} \cite{press2021train} (See Appendix~\ref{app:random_shift}). 
\subsection{1D Task: Language Modeling}
\label{sec:exp_lm}

\myparagraph{Setup} We first evaluate the performance of \textsc{SeqPE} on the language modeling task, where $\mathcal{L}_{main}$ is the cross entropy of the next-token prediction distribution. We follow the setup in \cite{press2021train} to train a language model for 100K steps with sequences of $L=512$ tokens on the training data of Wikitext-103, and evaluate it on the test set with extended sequences with $L\in[1024,16384]$ tokens. We use \texttt{huggingface/transformers}\footnote{\url{https://github.com/huggingface/transformers}} as our code-base and GPT-2 model \cite{radford2019language} as architecture for the main task. The best integration method of \textsc{SeqPE} is \textsc{AttnScalar} on the validation set. In this task, we share the parameters of $\mathbf{W}_{PE}^{q,k}$ in Equation~\ref{eq:pe_qk_map} for all layers of the GPT-2 model.  We compare \textsc{SeqPE} with strong position encoding baselines under the same setting:
\begin{itemize}
\item \textbf{RoPE}: Rotary positional encoding~\cite{su2024roformer} is widely used in language models~\cite{touvron2023llama}. Previous studies find that its base value significantly impacts extrapolation~\cite{men2024base, wu2025scaling}. Therefore, we train three variants with base values $10^4$, $10^6$, and $10^8$.

\item \textbf{ALiBi}: \textsc{ALiBi} introduces hand-crafted scalar biases to attention weights~\cite{almazrouei2023falcon, press2021train}. Despite its simplicity, it performs well in language modeling and extrapolation.

\item \textbf{APE}: Absolute position encoding sums token and position embeddings. We compare two variants: \textsc{APE-Sin} with fixed sinusoidal embeddings~\cite{vaswani2017attention}, and \textsc{APE-Learn} with learned embeddings~\cite{devlin2019bert, radford2019language}.

\item \textbf{NOPE}: Recent work shows that causal attention can be effective without PEs, referred to as \textsc{NOPE}~\cite{kazemnejad2023impact, yang2025RoPE}.
\end{itemize}
When testing on longer contexts, most methods, except \textsc{APE-Learn} that uses linear interpolation, generate new embeddings for unseen positions while keeping existing ones unchanged. More explanation for implementation details are in Appendix \ref{app:task1_lm_impl}.

\myparagraph{Results} As shown in Table \ref{tab:wt103}, our \textsc{SeqPE} demonstrates a significant advantage in the average score, indicating stronger generalization capability on Wikitext-103 test set. The base value of \textsc{RoPE} affects its final performance, but its extrapolation performance is still limited compared with \textsc{SeqPE} and \textsc{ALiBi}.

\begin{table*}[]
\centering
\caption{Accuracy ($\uparrow$) on ImageNet-1k. Results of \textsc{APE-Learn}, \textsc{RoPE2D},  \textsc{RoPE2D-Mix} are reported from the model pre-trained by \cite{heo2024rotary}. The training image resolution is $224\times 224$ with patch size $16\times 16$, while the testing resolutions are from $224\times 224$ to $672\times 672$. The best and second best results are represented by \textbf{bold text} and \underline{underline}, respectively.\label{tab:imgnet1k}}
\resizebox{0.7\textwidth}{!}{
\begin{tabular}{lcccccccc}\toprule
\textbf{Model}  & \textbf{224}          & \textbf{320}   & \textbf{384}    & \textbf{448}   & \textbf{512} & \textbf{640} & \textbf{672} & \textbf{AVG.}           \\\midrule
\textsc{NOPE}      &     78.8 &	79.2 &	78.5 &	77.3 &	75.8 &	\underline{71.7} &	\underline{70.5} &   75.9                             \\

\textsc{APE-Sin}      &   \underline{80.9} & 50.3 & 33.3 &	32.4 & 15.2 &	8.8 & 8.5   & 32.7                           \\\hdashline
\textsc{APE-Learn}      &   80.4    &     80.6     &    79.4   &    77.6      &    75.4   & 70.3 &   69.1     & 76.1                            \\
\textsc{RoPE2D}       &    \underline{80.9}    &  81.5      &  80.0     &   78.2     &  76.1   &  67.8 &  65.0    & 75.6                            \\
\textsc{RoPE2D-Mix}       &   \underline{80.9}     &   \textbf{82.2}    &   \textbf{81.8}   &    80.9    &   \underline{79.1}   & 71.6  &  68.3   & \underline{77.8}                             \\\midrule
\textsc{SeqPE}       &   \textbf{81.2} & 	\underline{81.8} &	\underline{81.7} &	\textbf{81.2} & \textbf{80.5} & \textbf{77.8} & \textbf{76.7} & 	\textbf{80.1}  \\\bottomrule  
\end{tabular}
}
\end{table*}

\subsection{1D Task: Fine-tuning for Long-context Question Answering}
\label{sec:qa}

\myparagraph{Setup} To assess the adaptability of various position encoding methods, we fine-tune the models from Section~\ref{sec:exp_lm} on a long-context QA task. Using the question answering (QA) prompt template in Fig.~\ref{fig:qa_prompt} (See Appendix) and the data pipeline from~\cite{hsieh2024ruler}, we generate 86,820 SQuAD-based QA pairs~\cite{rajpurkar-etal-2016-squad}, each with a multi-document context of approximately 1024 tokens. Models are fine-tuned for 10K steps using cross-entropy loss on the answer span, and we select the best model based on validation perplexity.

We evaluate on the long-context SQuAD dataset from the RULER benchmark~\cite{hsieh2024ruler} using two metrics. First, we compute perplexity of the gold answer span given the context and question. Then, for contexts ranging from 1024 to 8192 tokens, each model generates 16 tokens via greedy search, and we report the lower-cased Exact Match score against the gold answer. Additional implementation details are presented in Appendix~\ref{app:task2_qa_impl}.

\myparagraph{Results} As shown in Table \ref{tab:qa}, \textsc{SeqPE} achieves better average performance than both the \textsc{ALiBi} and \textsc{RoPE} baselines in terms of both perplexity and EM scores. Moreover, after switching to the QA downstream task, the extrapolation perplexity of \textsc{ALiBi} increases sharply when tested on sequences with 8K tokens, in contrast to its consistent performance on the pretraining task (See Section~\ref{sec:exp_lm}). In comparison, the perplexity of \textsc{SeqPE} remains relatively stable, demonstrating its strong adaptability.

\subsection{2D Task: Image Classification}
\label{sec:exp_img}

\myparagraph{Setup} One intriguing property of \textsc{SeqPE} is that it can be easily extended to data with higher dimensions using the sequential representation as a unified interface. Therefore, we further evaluate our method on the ImageNet-1K \cite{imagenet15russakovsky}, which contains $1.28$M training images and 50$K$ validation images of 1,000 classes. We follow official settings in \cite{DBLP:conf/icml/TouvronCDMSJ21} to train \textsc{ViT-S} models using different position encoding methods. 
The $\mathcal{L}_{main}$ is the cross entropy on the image classification distribution. The best integration method of \textsc{SeqPE} is \textsc{AttnMul} on the validation set. We compare \textsc{SeqPE} mainly with three groups of baselines:
\begin{itemize}[noitemsep, left=0pt, topsep=0pt]
    \item \textbf{APE}: The \textsc{APE-Learn} with learnable parameters is the PE strategy in \textsc{ViT} \cite{DBLP:conf/iclr/DosovitskiyB0WZ21}. We additionally evaluate the \textsc{APE-Sin} with expert-designed sinusoidal embeddings.
    \item \textbf{RoPE2D}: Since \textsc{RoPE} position encoding~\cite{su2024roformer} is proposed for 1-dim data,  Heo et al.~\cite{heo2024rotary} design a \textsc{RoPE2D} position encoding method for image data by considering the rotary operation for x-axis and y-axis separately. They also propose a variant \textsc{RoPE2D-Mix} with learnable frequencies. 
    \item \textbf{NOPE}: Following the experiments in Section~\ref{sec:exp_lm}, we also train a \textsc{ViT} without position encoding. 
\end{itemize}

We evaluate the extrapolation performance mainly following the setting in previous studies \cite{heo2024rotary, fan2024vitar}. In other words, we train the model with the $224\times224$ image resolution and testing on varied resolutions from $224\times224$ to $672\times672$. For all the image resolutions, each embedded patch represent $16\times 16$ pixels.

\begin{table}[]
\centering 
\caption{Ablation study for regularization objectives. We vary the values of $\alpha$ and $\beta$ for $\mathcal{L}_{\delta}$ and $\mathcal{L}_{OOD}$, respectively, while keeping the rest settings as same as those in Section~\ref{sec:exp_lm} and Section~\ref{sec:exp_img} for text and image tasks.
\label{tab:abalation_reg}}
\resizebox{1.0\columnwidth}{!}{
\begin{tabular}{lrrrrrr}\toprule
\multicolumn{7}{c}{\textit{Language Modeling} (\textit{Perplexity} $\downarrow$)}
\\\hline\hline
$\alpha$ & $\beta$ & \textbf{512}          & \textbf{1K}   & \textbf{2K}    & \textbf{4K} & \textbf{AVG.}           \\\hline\hline
$0.0$ & $0.0$ & \textbf{19.57} & 23.57 & 60.51 & 125.32 & 57.24 \\
$0.1$ & $0.0$ & \underline{19.62} & 18.78 & \underline{18.56} & \underline{18.62} & \underline{18.89} \\
$0.0$ & $0.1$ & \textbf{19.57} & 33.81 & 104.07 & 205.33 & 90.69 \\
$0.1$ & $0.1$       &    19.65 &	\textbf{18.74} &	\textbf{18.51} & 	\textbf{18.58}  & \textbf{18.86} \\
\midrule
\multicolumn{7}{c}{\textit{Image Classification} (\textit{Accuracy $\uparrow$})}
\\\hline\hline
& &  \textbf{224}          & \textbf{320}   & \textbf{384}    & \textbf{448} & \textbf{AVG.}           \\\hline\hline
$0.0$ & $0.0$ & 81.1 & 81.2 & 79.7 & 77.1 & 79.7 \\
$0.1$ & $0.0$ & 81.0 & 81.2 & \underline{80.7} & \underline{80.0} & \underline{80.7} \\
$0.0$ & $0.1$ & \textbf{81.3} & \underline{81.3} & 79.3 & 76.1 & 79.5 \\
$0.1$ & $0.1$ & \underline{81.2} & \textbf{81.8} & \textbf{81.7} & \textbf{80.5} & \textbf{81.3} \\\bottomrule  
\end{tabular}
}
\end{table}

\myparagraph{Results} As shown in Table \ref{tab:imgnet1k}, \textsc{SeqPE} significantly outperform all baselines in terms of averaged accuracy for different resolutions. It is also interesting to find that the \textsc{NOPE} baseline achieves relatively stable performance when testing on larger resolutions. 
We note that \textsc{RoPE2D-Mix} exhibits superior performance with slight increases in input resolutions (\textit{e.g.}, 320 and 384), yet it fails to maintain these advantages at larger resolutions. As a comparison, our \textsc{SeqPE} achieves consistent performance across various resolutions, demonstrating the superior generalization capability.

\subsection{Ablation Study}
\label{subsec:ablation}

We first demonstrate the effect of the two regularization objectives. As shown in Table \ref{tab:abalation_reg}, enabling the distance regularization objective $\mathcal{L}_{\delta}$ alone (\textit{i.e.}, $\alpha=0.1$, $\beta=0.0$) noticeably improves extrapolation performance on both the language modeling and image classification tasks. When the out-of-distribution objective $\mathcal{L}_{OOD}$ is used together with $\mathcal{L}_{\delta}$ (\textit{i.e.}, $\alpha=0.1$, $\beta=0.1$), the overall performance is further enhanced across both tasks. Interestingly, however, enabling $\mathcal{L}_{OOD}$ alone (\textit{i.e.}, $\alpha=0.0$, $\beta=0.1$) significantly degrades extrapolation performance, likely due to the potential embedding collapse discussed in Section~\ref{sec:pos_transferbility}.

In Table \ref{tab:ablation}, we compare different integration methods under the image classification setting described in Section~\ref{sec:exp_img}. We find that both \textsc{AttnMul} and \textsc{AttnBias} achieve strong extrapolation performance, with \textsc{AttnMul} performing slightly better. In contrast, \textsc{AttnSum} underperforms relative to the other methods, likely due to redundant correlation modeling between content and position embeddings \cite{kerethinking, su2024roformer}.

\begin{figure*}[t]
    \centering
    \begin{minipage}[b]{\textwidth}
        \centering
        \begin{minipage}[b]{0.19\textwidth}
            \includegraphics[width=\linewidth]{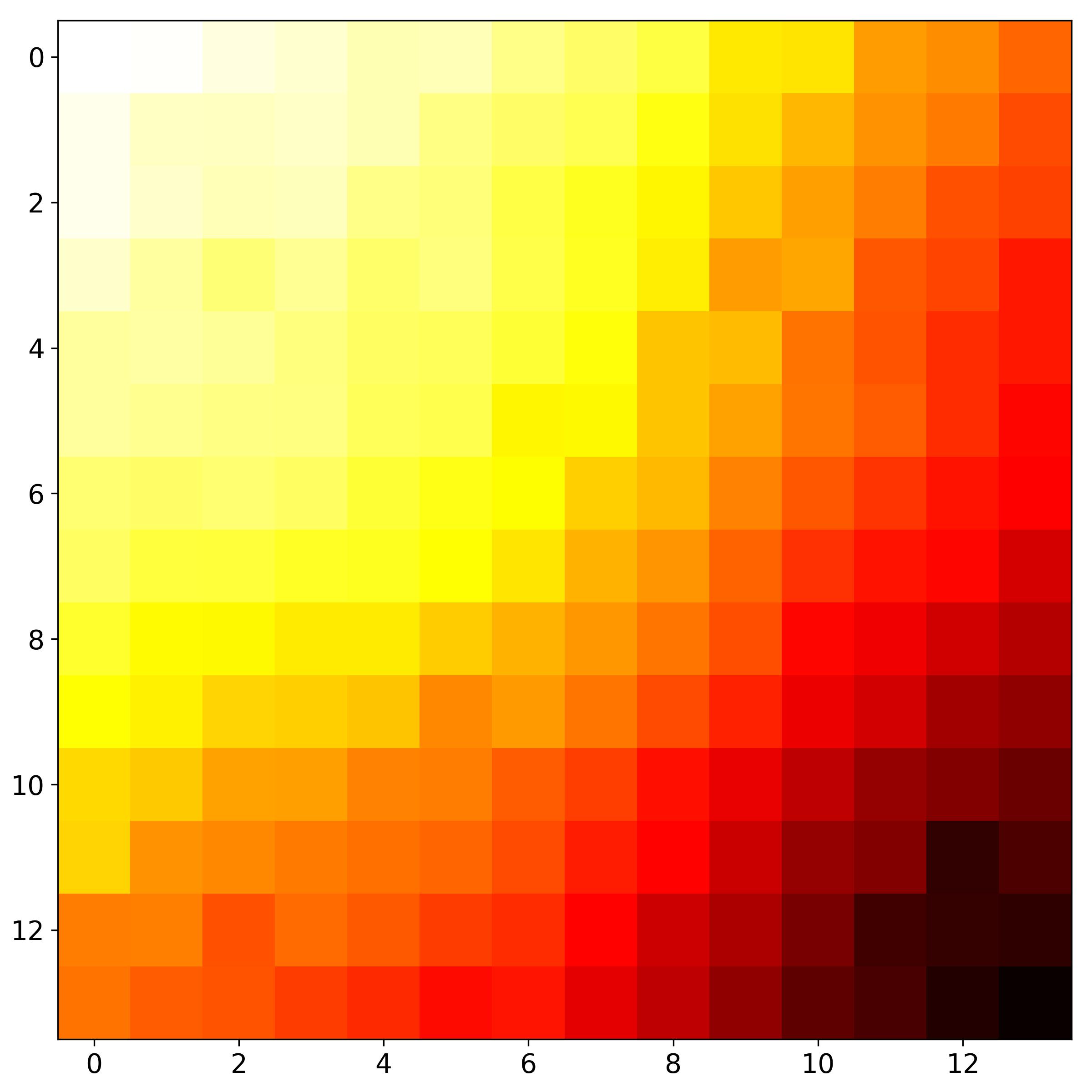}
            \caption*{\textsc{SeqPE} $(0,0)$}
        \end{minipage}
        \begin{minipage}[b]{0.19\textwidth}
            \includegraphics[width=\linewidth]{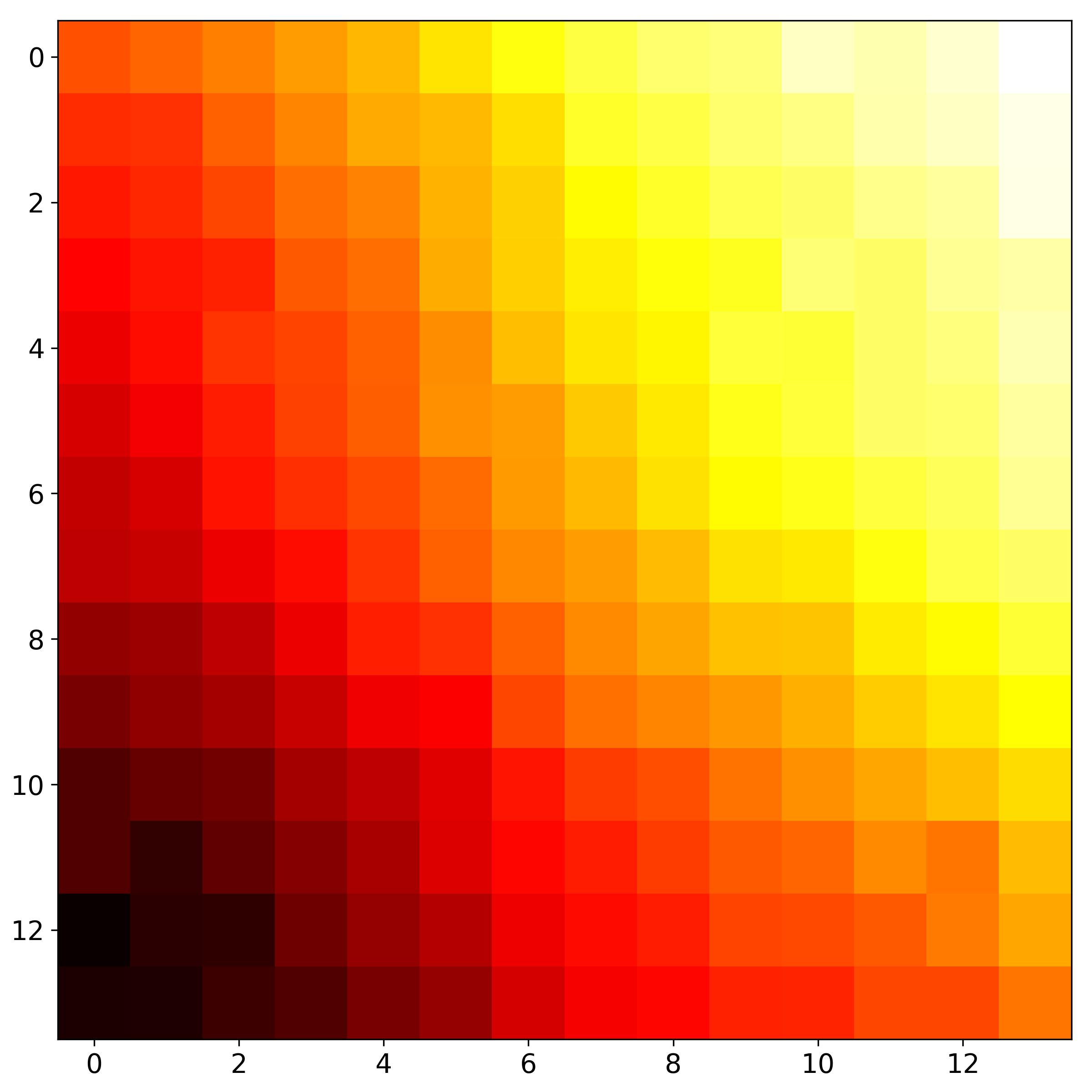}
            \caption*{\textsc{SeqPE} $(0,13)$}
        \end{minipage}
        \begin{minipage}[b]{0.19\textwidth}
            \includegraphics[width=\linewidth]{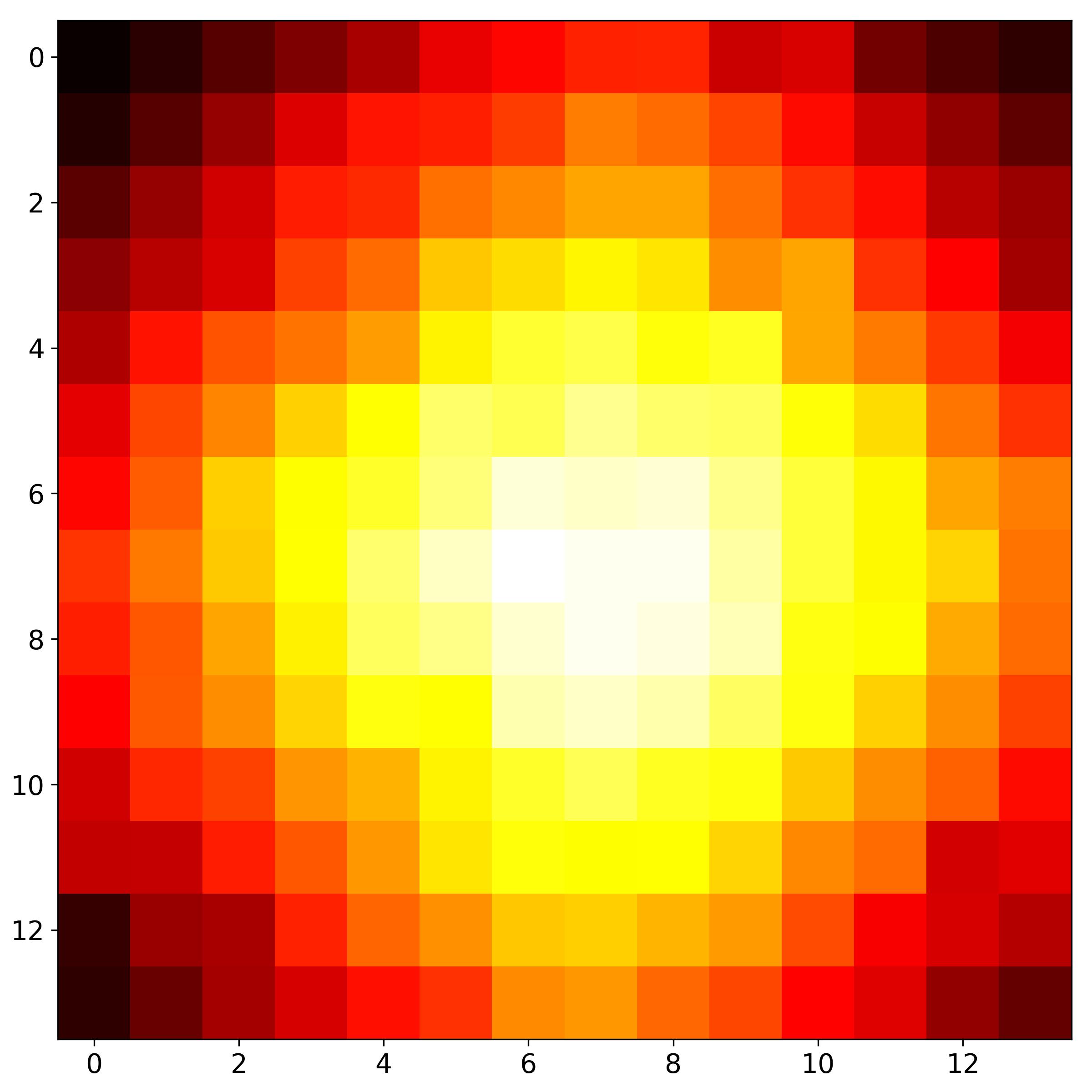}
            \caption*{\textsc{SeqPE} $(7,7)$}
        \end{minipage}
        \begin{minipage}[b]{0.19\textwidth}
            \includegraphics[width=\linewidth]{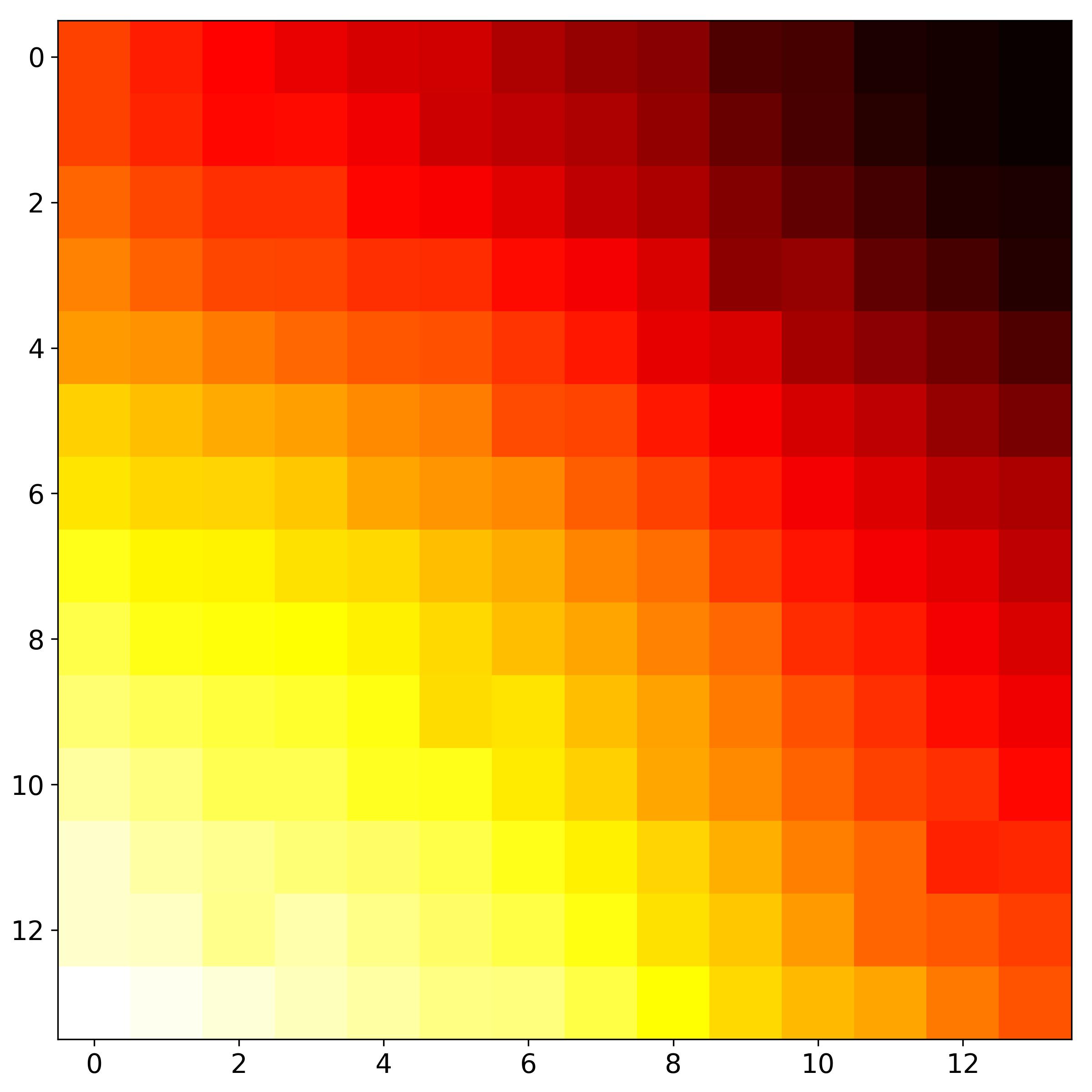}
            \caption*{\textsc{SeqPE} $(13,0)$}
        \end{minipage}
        \begin{minipage}[b]{0.19\textwidth}
            \includegraphics[width=\linewidth]{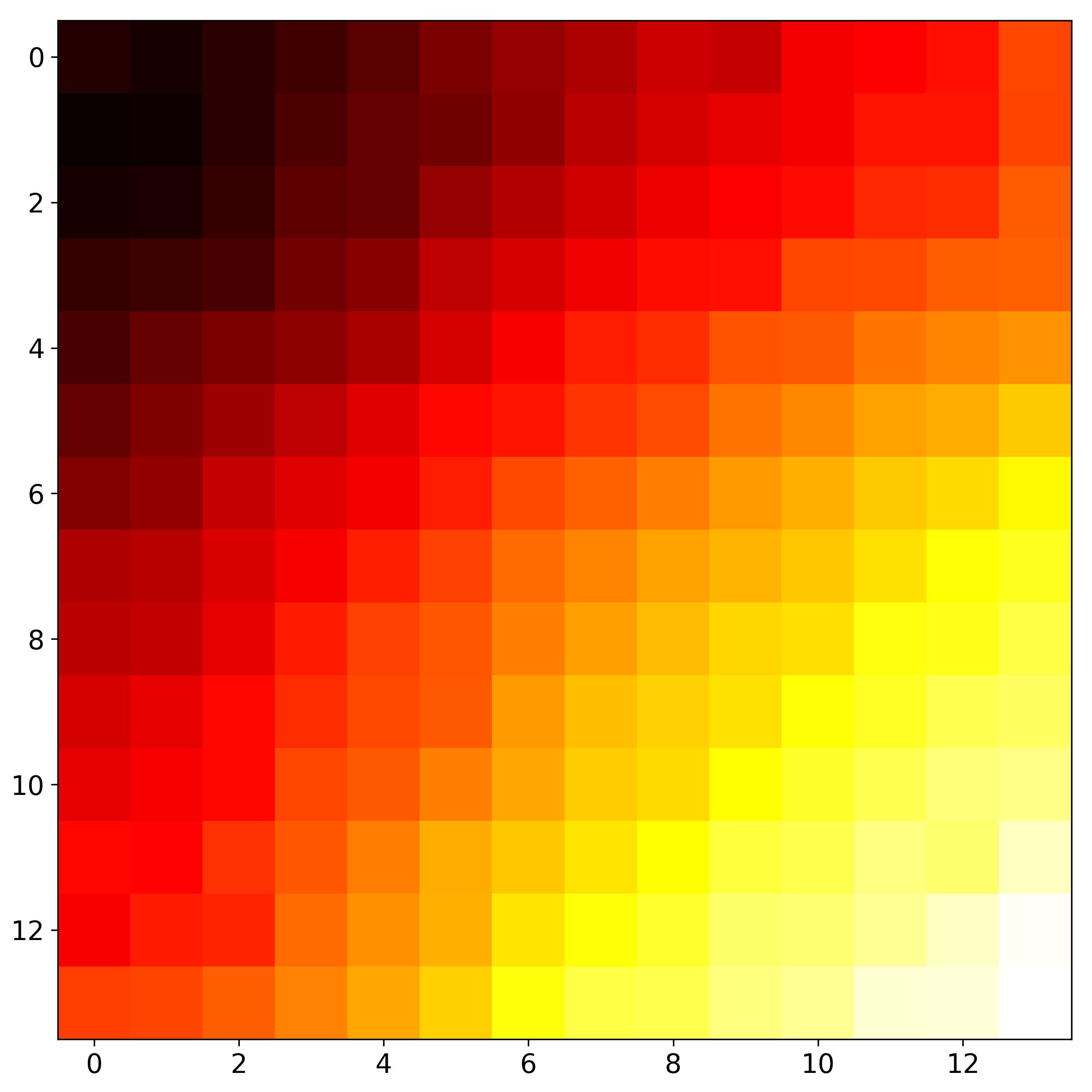}
            \caption*{\textsc{SeqPE} $(13,13)$}
        \end{minipage}
    \end{minipage}

    \vspace{1em}

    \begin{minipage}[b]{\textwidth}
        \centering
        \begin{minipage}[b]{0.19\textwidth}
            \includegraphics[width=\linewidth]{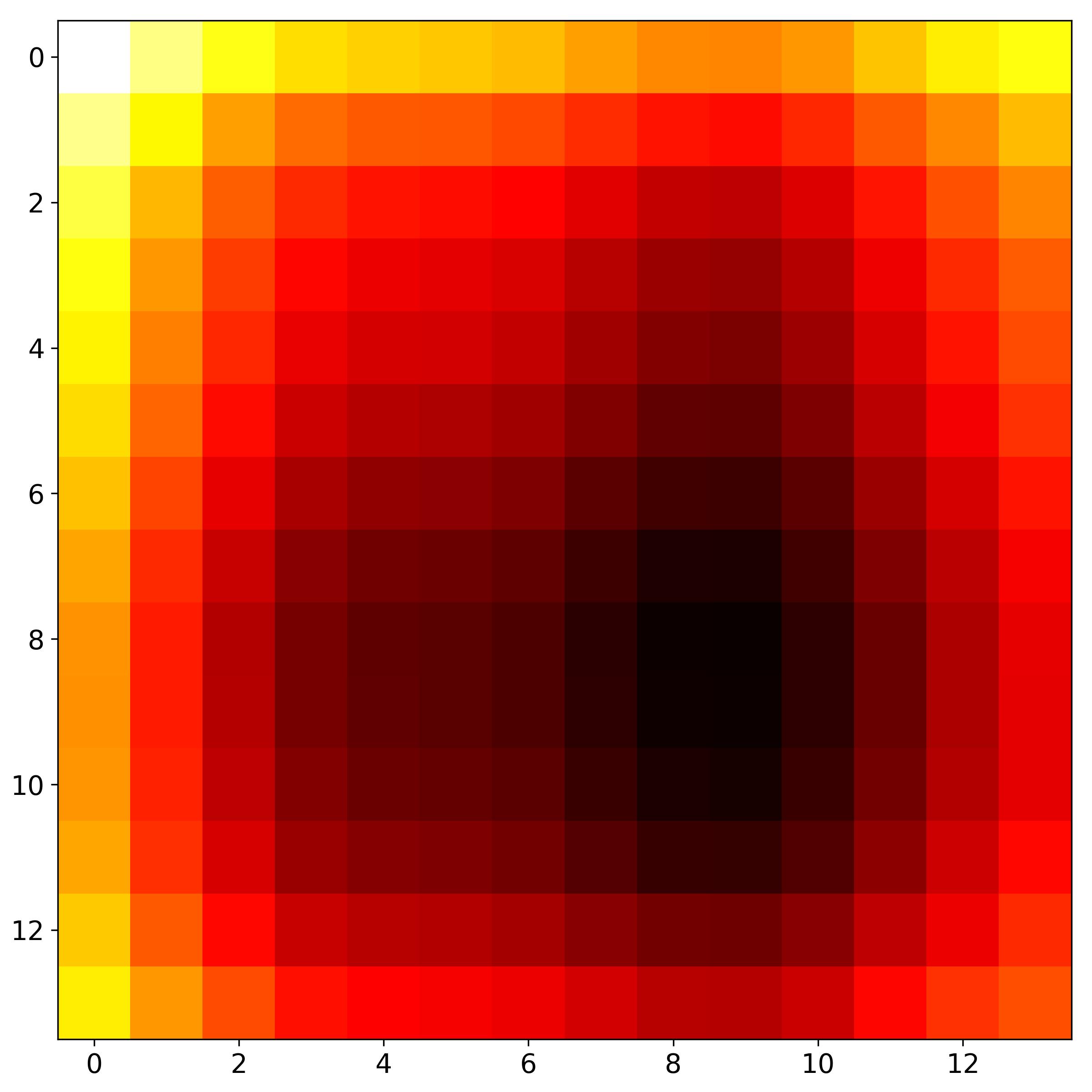}
            \caption*{\textsc{APE-Learn} $(0,0)$}
        \end{minipage}
        \begin{minipage}[b]{0.19\textwidth}
            \includegraphics[width=\linewidth]{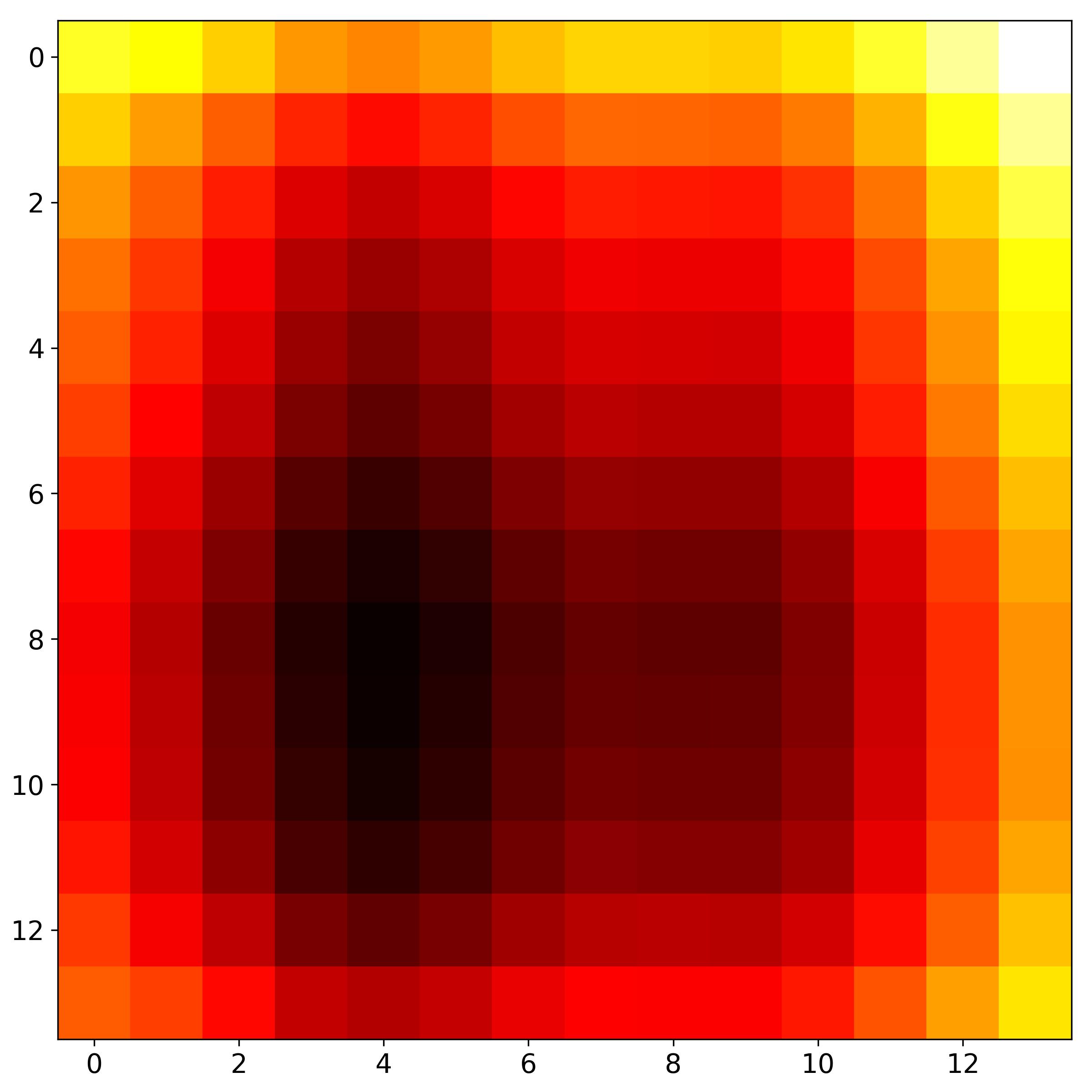}
            \caption*{\textsc{APE-Learn} $(0,13)$}
        \end{minipage}
        \begin{minipage}[b]{0.19\textwidth}
            \includegraphics[width=\linewidth]{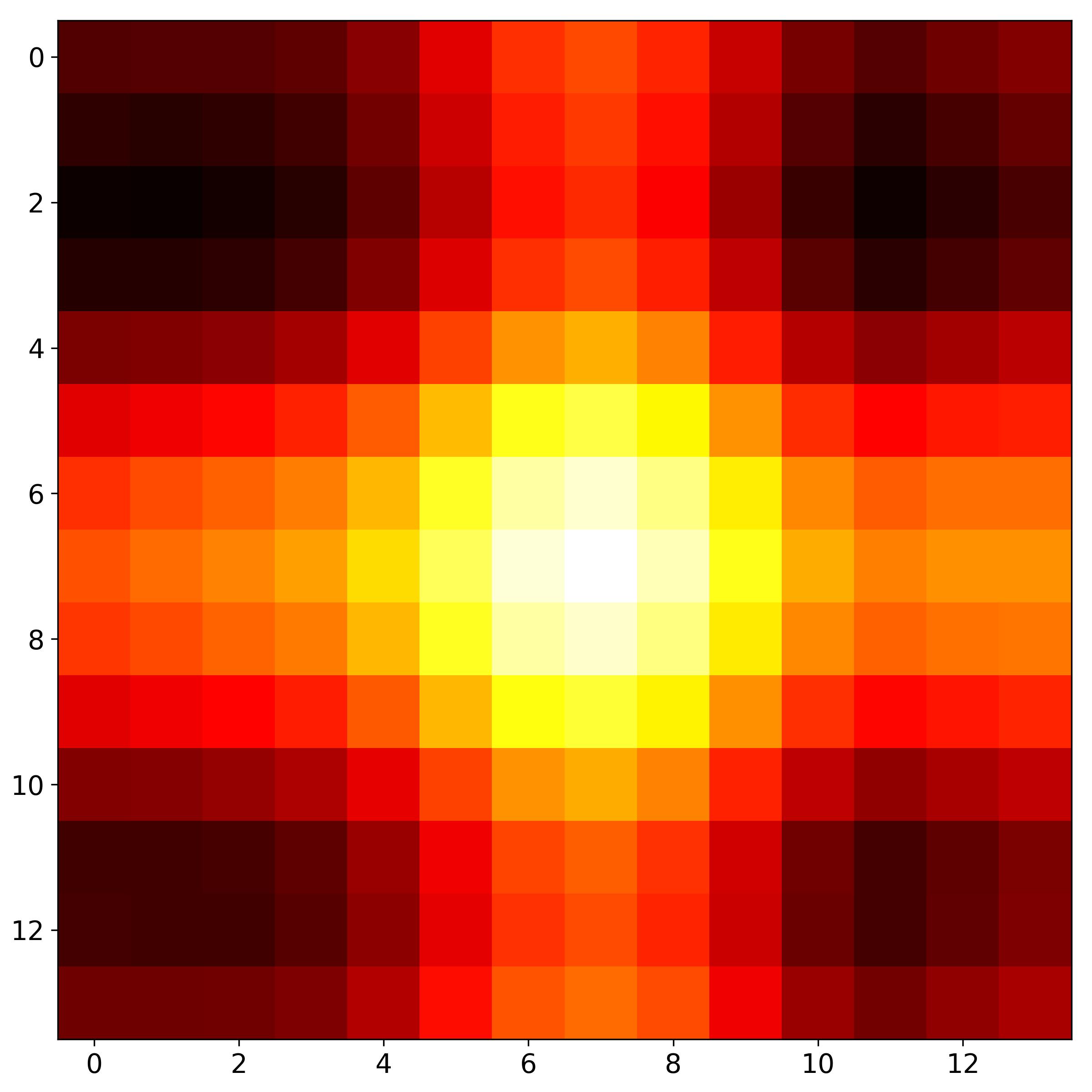}
            \caption*{\textsc{APE-Learn} $(7,7)$}
        \end{minipage}
        \begin{minipage}[b]{0.19\textwidth}
            \includegraphics[width=\linewidth]{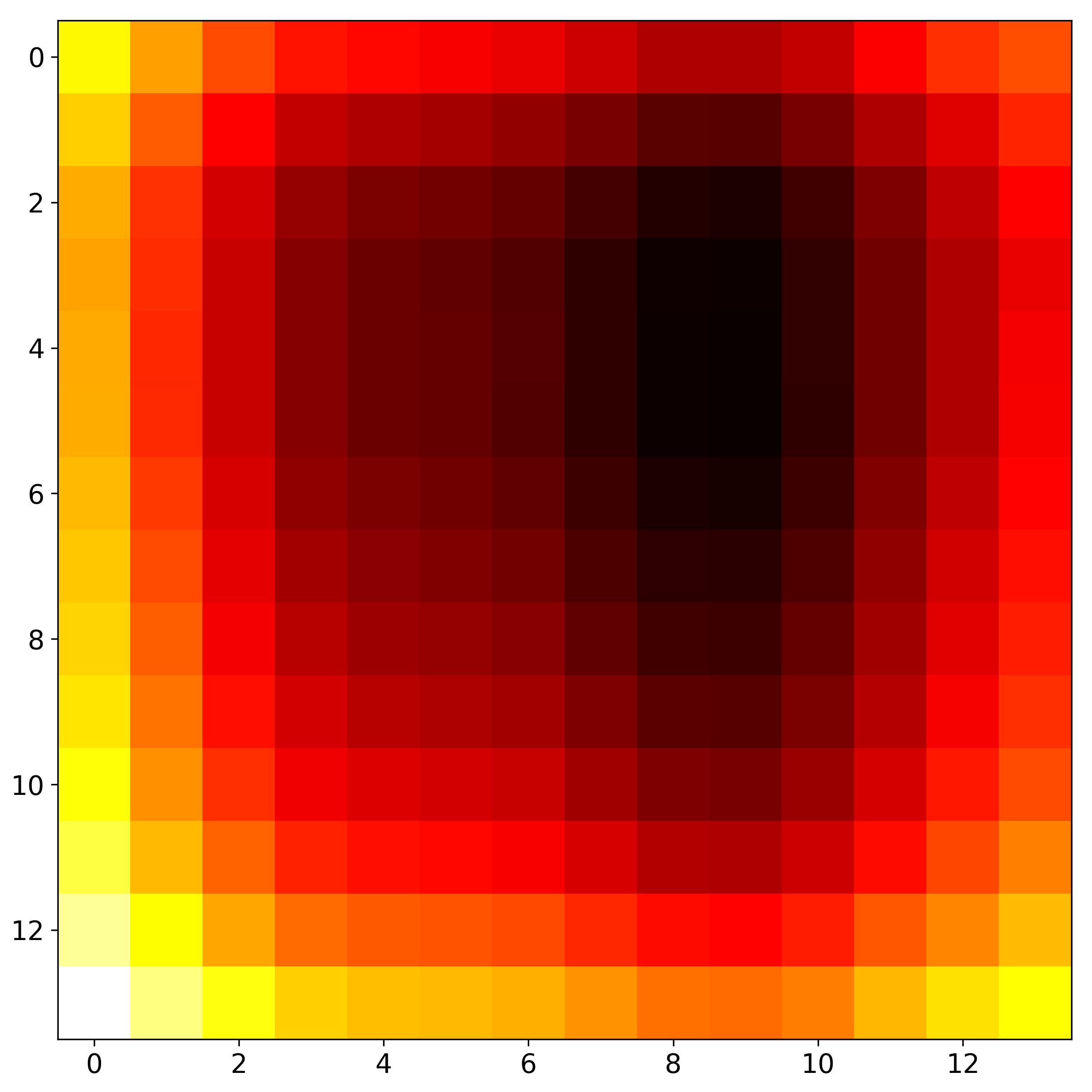}
            \caption*{\textsc{APE-Learn} $(13,0)$}
        \end{minipage}
        \begin{minipage}[b]{0.19\textwidth}
            \includegraphics[width=\linewidth]{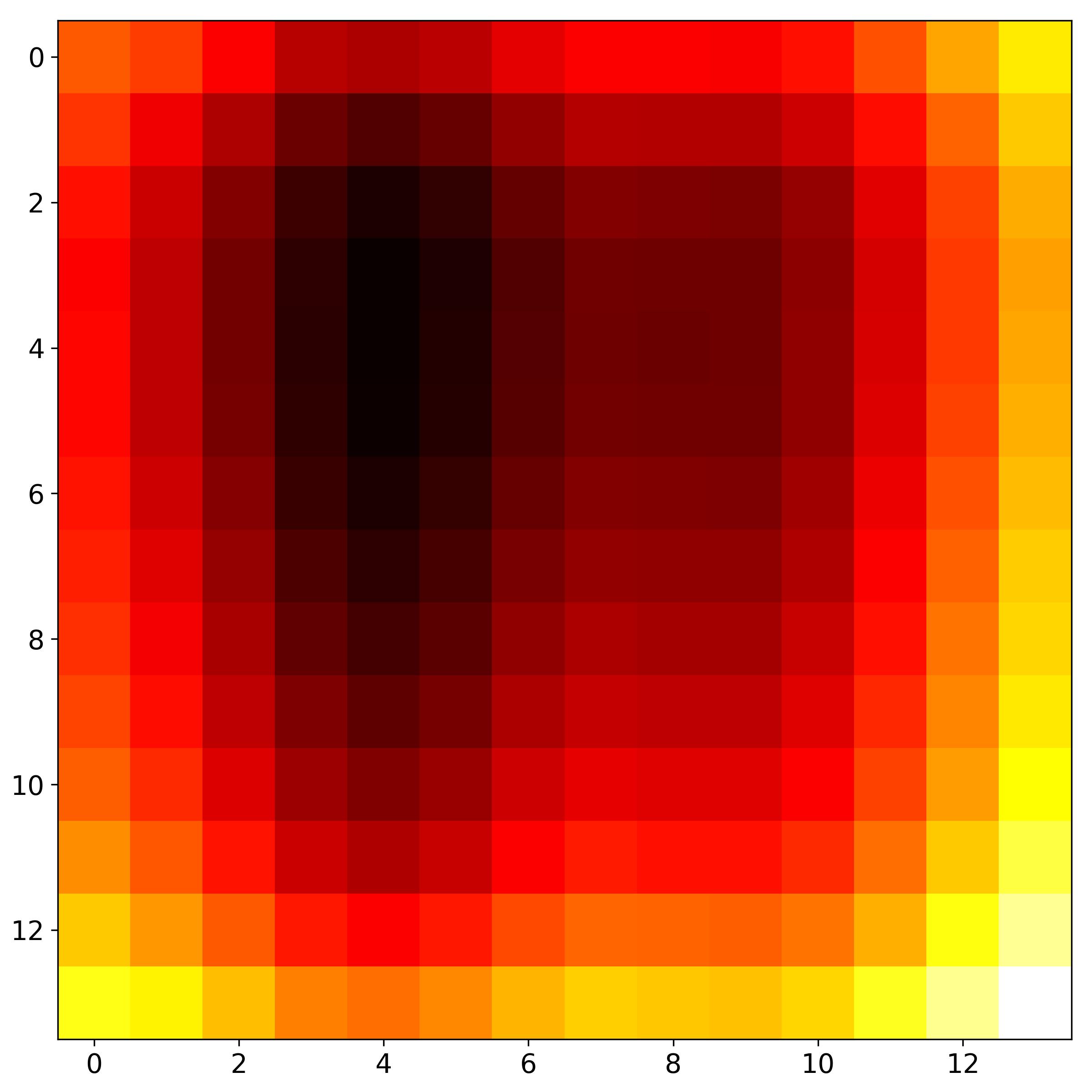}
            \caption*{\textsc{APE-Learn} $(13,13)$}
        \end{minipage}
    \end{minipage}
    \caption{Heatmap for position embeddings of 2D images with $14\times14$ patches. For each sub-figure, we plot the heatmap based on the 5 selected positions. Brighter area indices larger dot-product values.}
    \label{fig:case_study}
\end{figure*}

Additionally, in Fig.~\ref{fig:case_study}, we visualize the relationship between position embeddings (PEs) obtained from learnable position encoding methods, \emph{i.e.}, \textsc{SeqPE} and \textsc{APE-Learn}, trained under the 2D positional setting described in Section~\ref{sec:exp_img}. Specifically, we select five positions, four at the corners and one at the center, and compute the dot product between each of these selected positions and all other positions in the 2D image. Compared to the patterns produced by the APE method, we observe that \textsc{SeqPE} better captures the relative relationships between positions.

\begin{table}[t]
\centering  
\caption{Ablation study for integration methods on the image classification task. We only vary the integration methods while keeping rest settings the same as those in Section~\ref{sec:exp_img}. Accuracy ($\uparrow$) is reported. \label{tab:ablation}}
\resizebox{1.0\linewidth}{!}{
\begin{tabular}{cccccccccc}\toprule
  Integration & \textbf{224}          & \textbf{320}   & \textbf{384}    & \textbf{448}   &  \textbf{AVG.}           \\\midrule
 \textsc{AttnSum} & 80.9 & 80.8 & 79.8 & 78.5 & 80.0 \\
 \textsc{AttnBias} & 80.8 & 80.9 & 80.1 & 79.2 & 80.2 \\
 \textsc{AttnMul} & 81.2 & 81.8 & 81.7 & 81.2 & 81.4 \\
\bottomrule           
\end{tabular}
}
\end{table}

\section{Related Work}
\label{sec:related_work}

\subsection{Position Encoding Schemes} 
In Transformer-based methods, the foundational work in position encoding includes Sinusoidal Encoding~\cite{vaswani2017attention}, which uses trigonometric functions to encode absolute positions while capturing relative relationships. 
Subsequent research has explored various variants of encoding positional information. 

From the perspective of input, it is common that the positional information is parameterized by employing a lookup table. These methods can be broadly categorized into absolute position embeddings (APE)~\cite{vaswani2017attention,devlin2019bert,wang2019encoding,DBLP:conf/iclr/DosovitskiyB0WZ21,ren2023masked} (\textit{i.e.}, assigning unique embeddings to each absolute position) and relative position embeddings (RPE)~\cite{huang2020improve,raffel2020exploring} (\textit{i.e.}, explicitly modeling relative distances between tokens). 
Specifically, for some methods~\cite{vaswani2017attention}, the parameters of PEs are fixed during model training, while others~\cite{devlin2019bert,liu2021swin} are learnable.
However, most these methods are either trained on fixed-length sequences or lack inherent inductive biases for positional relationships (\textit{e.g.}, translation invariance), leading to poor extrapolation to longer sequences during inference.

From the perspective of attention mechanism,  
injecting positional information by modifying attention scores with predefined learnable biases is also a promising direction.
For example, 
Shaw et al.~\cite{shaw2018self} first introduce learnable embeddings for relative positions that are used to modify key and value vectors, or directly added to attention logits. 
T5~\cite{raffel2020exploring} bias method learns a scalar bias for each relative distance bin, added to attention scores before softmax. \textsc{ALiBi}~\cite{press2021train,almazrouei2023falcon} adds a linear penalty to attention scores based on the relative distance between tokens. 
While more complex, DeBERTa~\cite{he2020deberta} uses disentangled matrices to separately represent content and position, and the relative position information also influences the attention scores in a bias-like manner. 
Subsequent, Swin Transformer~\cite{liu2021swin}
creates a learnable bias parameter table where the table size is determined by the possible range of relative coordinates within local windows. 
Notably, Rotary Position Embedding (RoPE)~\cite{su2024roformer} emerges as popular alternatives, with RoPE integrating relative positional information through rotational transformation on the query and key vectors. LiePE~\cite{ostmeier2024liere} generalizes
RoPE to high-dimensional rotation matrices by
leveraging Lie group theory, which replaces RoPE’s rotation matrices with learned, dense rotations derived from Lie group generators. 
Although many of those PEs offer significantly advantages in terms of extrapolation, efficiency and capturing relative positional information, they also come with several limitations: (1) They primarily  model ``distance decay'' or ``local emphasis'' but may fail to capture more subtle, non-monotonic or directional positional relationships that learnable embeddings could represent. (2) Methods like \textsc{ALiBi} use a fixed, non-learned functional form for the bias lack adaptability for different datasets, tasks~\cite{li2024aria,zhou2024transfusion} or even different layers within the same model~\cite{haviv2022transformer}. We have proved that SeqPE can significantly improve on these limitations.

Among these aforementioned methods, 
\textsc{FLOATER}~\cite{liu2020learning} is the most relevant work to ours, which used the Neural ODEs~\cite{chen2018neural} to generate position embeddings that are integrated into each Transformer block.
In comparison, our proposed SeqPE has three distinct advantages: (1) Comparing to the $O(n)$ time complexity of \textsc{FLOATER}, our position encoder is a lightweight module that barely increases the training and inference complexities. 
(2) We use explicit position rules to constrain PEs, offering better interpretability than the ``black-box'' ODEs. (3) Unlike the carefully tuned layer-specific initial values in FLOATER, our method is insensitive to the initialization values. 
Moreover, we demonstrate that the proposed SeqPE is a unified framework that can be applied to both the input embeddings and the attention mechanism.

\subsection{Advances in Long-Context Modeling}
Recently, employing the large language model (LLM) to process text beyond their training sequence length or image beyond their training resolution, \textit{i.e.}, extrapolation, has become a prominent research topic. 
Many research works attribute the poor extrapolate of LLMs to the position encoding~\cite{press2021train, zhao2023length}. 
For example, Fourier Position Embedding (FoPE)~\cite{hua2024fourier} enhances the frequency-domain properties of attention to increase the model's robustness against the spectrum damage, helping in better handling of long-range dependencies in sequence. 
Recently,
CLEX~\cite{chen2024clex} introduces a continuous scaling framework based on ODEs to generalize positional embedding adjustments dynamically for \textsc{RoPE}, allowing LLMs to extrapolate context windows 4–8$\times$ beyond their training lengths. However, since these methods~\cite{chen2023extending, pengyarn, chen2024clex} rely on a pre-specified context length to determine scaling factors, they may not generalize well to fractional or arbitrarily stretched context lengths, impairing the model's understanding of token order. Moreover, in generative models~\cite{touvron2023llama, yang2025qwen3, guo2025deepseek}, where the output sequence length is unknown in advance, such dependence can introduce challenges during inference, especially when using key-value caching in Transformer models~\cite{kwon2023efficient}.
In contrast to prior approaches, we propose a knowledge distillation method that smoothly bridges the gap between positions within and beyond the training context length, preserving consistent positional embeddings during extrapolation across variable-length sequences.  

Some methods~\cite{ruoss-etal-2023-randomized,zhu2023pose,fan2024vitar} improve the generalization of model on unseen positions by introducing randomness to PEs during training, preventing overfitting to specific context length or image resolitions. 
Similarly, we also introduce a random shift for the starting position for a better generalization in our proposed method.

\subsection{Contrastive Learning in PEs.} Contrastive learning is a powerful technique used to learn robust and meaningful embeddings of data. The goal is to map data into an embedding space where similar items are close and dissimilar ones are far apart, which has achieved remarkable performance on various tasks~\cite{chen2020simple,he2020momentum,grill2020bootstrap,gao-etal-2021-simcse,radford2021learning}. As far as we know, we are the first to propose that when learning PEs, the sequential or spatial relationships between PEs are reinforced in a self-supervised manner by introducing contrastive learning.

\section{Conclusion}
\label{sec:conclusion}
In this work, we introduce a lightweight sequential position encoder to learn position embeddings in Transformer-based models, referred to as sequential position encoding (\textsc{SeqPE}). To regularize the representation space of \textsc{SeqPE}, we propose two complementary objectives that aim to align embedding distances with a predefined position-distance function and enhance the generalization for out-of-distribution positions. On tasks involving one-dimensional data, \emph{i.e.}, language modeling and question answering, our \textsc{SeqPE}, jointly trained with the main task model in an end-to-end manner, demonstrates greater adaptability on downstream tasks and achieves stronger extrapolation performance. Furthermore, when transforming \textsc{SeqPE} from 1-dimensional tasks to the 2-dimensional image classification, \textsc{SeqPE} outperforms strong baselines such as \textsc{RoPE2D} by a large margin, with almost no manual architectural redesign. 
These experimental results validate that our method demonstrates superior learnability and adaptability across diverse tasks, exhibits robust context extrapolation capabilities that enable effective generalization to sequences significantly longer than training examples, and offers seamless extensibility to arbitrary-dimensional data with minimal manual intervention.


\bibliography{custom}

\begin{thebibliography}{10}

\bibitem{almazrouei2023falcon}
Ebtesam Almazrouei, Hamza Alobeidli, Abdulaziz Alshamsi, Alessandro Cappelli, Ruxandra Cojocaru, M{\'e}rouane Debbah, {\'E}tienne Goffinet, Daniel Hesslow, Julien Launay, Quentin Malartic, et~al.
\newblock The falcon series of open language models.
\newblock {\em arXiv preprint arXiv:2311.16867}, 2023.

\bibitem{bandel-etal-2022-lexical}
Elron Bandel, Yoav Goldberg, and Yanai Elazar.
\newblock Lexical generalization improves with larger models and longer training.
\newblock In Yoav Goldberg, Zornitsa Kozareva, and Yue Zhang, editors, {\em Findings of the Association for Computational Linguistics: EMNLP}, 2022.

\bibitem{brown2020language}
Tom Brown, Benjamin Mann, Nick Ryder, Melanie Subbiah, Jared~D Kaplan, Prafulla Dhariwal, Arvind Neelakantan, Pranav Shyam, Girish Sastry, Amanda Askell, et~al.
\newblock Language models are few-shot learners.
\newblock {\em Advances in Neural Information Processing Systems (NeurIPS)}, 2020.

\bibitem{chen2024clex}
Guanzheng Chen, Xin Li, Zaiqiao Meng, Shangsong Liang, and Lidong Bing.
\newblock {CLEX}: Continuous length extrapolation for large language models.
\newblock In {\em International Conference on Learning Representations (ICLR)}, 2024.

\bibitem{chen2018neural}
Ricky~TQ Chen, Yulia Rubanova, Jesse Bettencourt, and David~K Duvenaud.
\newblock Neural ordinary differential equations.
\newblock {\em Advances in Neural Information Processing Systems (NeurIPS)}, 2018.

\bibitem{chen2023extending}
Shouyuan Chen, Sherman Wong, Liangjian Chen, and Yuandong Tian.
\newblock Extending context window of large language models via positional interpolation.
\newblock {\em arXiv preprint arXiv:2306.15595}, 2023.

\bibitem{chen2020simple}
Ting Chen, Simon Kornblith, Mohammad Norouzi, and Geoffrey Hinton.
\newblock A simple framework for contrastive learning of visual representations.
\newblock In {\em International Conference on Machine Learning (ICML)}, 2020.

\bibitem{devlin2019bert}
Jacob Devlin, Ming-Wei Chang, Kenton Lee, and Kristina Toutanova.
\newblock Bert: Pre-training of deep bidirectional transformers for language understanding.
\newblock In {\em Proceedings of the conference of the North American Chapter of the Association for Computational Linguistics (NAACL)}, 2019.

\bibitem{DBLP:conf/iclr/DosovitskiyB0WZ21}
Alexey Dosovitskiy, Lucas Beyer, Alexander Kolesnikov, Dirk Weissenborn, Xiaohua Zhai, Thomas Unterthiner, Mostafa Dehghani, Matthias Minderer, Georg Heigold, Sylvain Gelly, Jakob Uszkoreit, and Neil Houlsby.
\newblock An image is worth 16x16 words: Transformers for image recognition at scale.
\newblock In {\em International Conference on Learning Representations (ICLR)}, 2021.

\bibitem{fan2024vitar}
Qihang Fan, Quanzeng You, Xiaotian Han, Yongfei Liu, Yunzhe Tao, Huaibo Huang, Ran He, and Hongxia Yang.
\newblock Vitar: Vision transformer with any resolution.
\newblock {\em arXiv preprint arXiv:2403.18361}, 2024.

\bibitem{gao-etal-2021-simcse}
Tianyu Gao, Xingcheng Yao, and Danqi Chen.
\newblock {S}im{CSE}: Simple contrastive learning of sentence embeddings.
\newblock In Marie-Francine Moens, Xuanjing Huang, Lucia Specia, and Scott Wen-tau Yih, editors, {\em Proceedings of the Conference on Empirical Methods in Natural Language Processing (EMNLP)}, 2021.

\bibitem{gehring2017convolutional}
Jonas Gehring, Michael Auli, David Grangier, Denis Yarats, and Yann~N Dauphin.
\newblock Convolutional sequence to sequence learning.
\newblock In {\em International Conference on Machine Learning (ICML)}, pages 1243--1252, 2017.

\bibitem{girdhar2023imagebind}
Rohit Girdhar, Alaaeldin El-Nouby, Zhuang Liu, Mannat Singh, Kalyan~Vasudev Alwala, Armand Joulin, and Ishan Misra.
\newblock Imagebind: One embedding space to bind them all.
\newblock In {\em Proceedings of the IEEE/CVF Conference on Computer Vision and Pattern Recognition (CVPR)}, 2023.

\bibitem{grill2020bootstrap}
Jean-Bastien Grill, Florian Strub, Florent Altch{\'e}, Corentin Tallec, Pierre Richemond, Elena Buchatskaya, Carl Doersch, Bernardo Avila~Pires, Zhaohan Guo, Mohammad Gheshlaghi~Azar, et~al.
\newblock Bootstrap your own latent-a new approach to self-supervised learning.
\newblock {\em Advances in Neural Information Processing Systems (NeurIPS)}, 2020.

\bibitem{guo2025deepseek}
Daya Guo, Dejian Yang, Haowei Zhang, Junxiao Song, Ruoyu Zhang, Runxin Xu, Qihao Zhu, Shirong Ma, Peiyi Wang, Xiao Bi, et~al.
\newblock Deepseek-r1: Incentivizing reasoning capability in llms via reinforcement learning.
\newblock {\em arXiv preprint arXiv:2501.12948}, 2025.

\bibitem{haviv2022transformer}
Adi Haviv, Ori Ram, Ofir Press, Peter Izsak, and Omer Levy.
\newblock Transformer language models without positional encodings still learn positional information.
\newblock {\em arXiv preprint arXiv:2203.16634}, 2022.

\bibitem{he2020momentum}
Kaiming He, Haoqi Fan, Yuxin Wu, Saining Xie, and Ross Girshick.
\newblock Momentum contrast for unsupervised visual representation learning.
\newblock In {\em Proceedings of the IEEE/CVF Conference on Computer Vision and Pattern Recognition}, 2020.

\bibitem{he2020deberta}
Pengcheng He, Xiaodong Liu, Jianfeng Gao, and Weizhu Chen.
\newblock Deberta: Decoding-enhanced bert with disentangled attention.
\newblock In {\em International Conference on Learning Representations (ICLR)}, 2020.

\bibitem{heo2024rotary}
Byeongho Heo, Song Park, Dongyoon Han, and Sangdoo Yun.
\newblock Rotary position embedding for vision transformer.
\newblock In {\em European Conference on Computer Vision (ECCV)}, 2024.

\bibitem{hsieh2024ruler}
Cheng-Ping Hsieh, Simeng Sun, Samuel Kriman, Shantanu Acharya, Dima Rekesh, Fei Jia, and Boris Ginsburg.
\newblock {RULER}: What{\textquoteright}s the real context size of your long-context language models?
\newblock In {\em First Conference on Language Modeling}, 2024.

\bibitem{hua2024fourier}
Ermo Hua, Che Jiang, Xingtai Lv, Kaiyan Zhang, Ning Ding, Youbang Sun, Biqing Qi, Yuchen Fan, Xuekai Zhu, and Bowen Zhou.
\newblock Fourier position embedding: Enhancing attention's periodic extension for length generalization.
\newblock {\em arXiv preprint arXiv:2412.17739}, 2024.

\bibitem{huang2020improve}
Zhiheng Huang, Davis Liang, Peng Xu, and Bing Xiang.
\newblock Improve transformer models with better relative position embeddings.
\newblock {\em arXiv preprint arXiv:2009.13658}, 2020.

\bibitem{jia-liang-2017-adversarial}
Robin Jia and Percy Liang.
\newblock Adversarial examples for evaluating reading comprehension systems.
\newblock In Martha Palmer, Rebecca Hwa, and Sebastian Riedel, editors, {\em Proceedings of the Conference on Empirical Methods in Natural Language Processing (EMNLP)}, 2017.

\bibitem{kazemnejad2023impact}
Amirhossein Kazemnejad, Inkit Padhi, Karthikeyan Natesan~Ramamurthy, Payel Das, and Siva Reddy.
\newblock The impact of positional encoding on length generalization in transformers.
\newblock {\em Advances in Neural Information Processing Systems (NeurIPS)}, 2023.

\bibitem{kerethinking}
Guolin Ke, Di~He, and Tie-Yan Liu.
\newblock Rethinking positional encoding in language pre-training.
\newblock In {\em International Conference on Learning Representations}.

\bibitem{kwon2023efficient}
Woosuk Kwon, Zhuohan Li, Siyuan Zhuang, Ying Sheng, Lianmin Zheng, Cody~Hao Yu, Joseph Gonzalez, Hao Zhang, and Ion Stoica.
\newblock Efficient memory management for large language model serving with pagedattention.
\newblock In {\em Proceedings of the 29th Symposium on Operating Systems Principles}, pages 611--626, 2023.

\bibitem{li2024aria}
Dongxu Li, Yudong Liu, Haoning Wu, Yue Wang, Zhiqi Shen, Bowen Qu, Xinyao Niu, Fan Zhou, Chengen Huang, Yanpeng Li, et~al.
\newblock Aria: An open multimodal native mixture-of-experts model.
\newblock {\em arXiv preprint arXiv:2410.05993}, 2024.

\bibitem{liu2020learning}
Xuanqing Liu, Hsiang-Fu Yu, Inderjit Dhillon, and Cho-Jui Hsieh.
\newblock Learning to encode position for transformer with continuous dynamical model.
\newblock In {\em International Conference on Machine Learning (ICML)}, 2020.

\bibitem{liu2021swin}
Ze~Liu, Yutong Lin, Yue Cao, Han Hu, Yixuan Wei, Zheng Zhang, Stephen Lin, and Baining Guo.
\newblock Swin transformer: Hierarchical vision transformer using shifted windows.
\newblock In {\em Proceedings of the IEEE/CVF International Conference on Computer Vision (ICCV)}, 2021.

\bibitem{men2024base}
Xin Men, Mingyu Xu, Bingning Wang, Qingyu Zhang, Hongyu Lin, Xianpei Han, and Weipeng Chen.
\newblock Base of rope bounds context length.
\newblock {\em arXiv preprint arXiv:2405.14591}, 2024.

\bibitem{ostmeier2024liere}
Sophie Ostmeier, Brian Axelrod, Michael~E Moseley, Akshay Chaudhari, and Curtis Langlotz.
\newblock Liere: Generalizing rotary position encodings.
\newblock {\em arXiv preprint arXiv:2406.10322}, 2024.

\bibitem{pengyarn}
Bowen Peng, Jeffrey Quesnelle, Honglu Fan, and Enrico Shippole.
\newblock Yarn: Efficient context window extension of large language models.
\newblock In {\em The Twelfth International Conference on Learning Representations}.

\bibitem{press2021train}
Ofir Press, Noah Smith, and Mike Lewis.
\newblock Train short, test long: Attention with linear biases enables input length extrapolation.
\newblock In {\em International Conference on Learning Representations (ICLR)}, 2021.

\bibitem{radford2021learning}
Alec Radford, Jong~Wook Kim, Chris Hallacy, Aditya Ramesh, Gabriel Goh, Sandhini Agarwal, Girish Sastry, Amanda Askell, Pamela Mishkin, Jack Clark, et~al.
\newblock Learning transferable visual models from natural language supervision.
\newblock In {\em International Conference on Machine Learning (ICML)}, 2021.

\bibitem{radford2019language}
Alec Radford, Jeffrey Wu, Rewon Child, David Luan, Dario Amodei, Ilya Sutskever, et~al.
\newblock Language models are unsupervised multitask learners.
\newblock {\em OpenAI blog}, 1(8):9, 2019.

\bibitem{raffel2020exploring}
Colin Raffel, Noam Shazeer, Adam Roberts, Katherine Lee, Sharan Narang, Michael Matena, Yanqi Zhou, Wei Li, and Peter~J Liu.
\newblock Exploring the limits of transfer learning with a unified text-to-text transformer.
\newblock {\em Journal of Machine Learning Research}, 21:1--67, 2020.

\bibitem{rajpurkar-etal-2016-squad}
Pranav Rajpurkar, Jian Zhang, Konstantin Lopyrev, and Percy Liang.
\newblock {SQ}u{AD}: 100,000+ questions for machine comprehension of text.
\newblock In Jian Su, Kevin Duh, and Xavier Carreras, editors, {\em Proceedings of the Conference on Empirical Methods in Natural Language Processing (EMNLP)}, 2016.

\bibitem{ren2023masked}
Bin Ren, Yahui Liu, Yue Song, Wei Bi, Rita Cucchiara, Nicu Sebe, and Wei Wang.
\newblock Masked jigsaw puzzle: A versatile position embedding for vision transformers.
\newblock In {\em Proceedings of the IEEE/CVF Conference on Computer Vision and Pattern Recognition (CVPR)}, 2023.

\bibitem{ruoss-etal-2023-randomized}
Anian Ruoss, Gr{\'e}goire Del{\'e}tang, Tim Genewein, Jordi Grau-Moya, R{\'o}bert Csord{\'a}s, Mehdi Bennani, Shane Legg, and Joel Veness.
\newblock Randomized positional encodings boost length generalization of transformers.
\newblock In Anna Rogers, Jordan Boyd-Graber, and Naoaki Okazaki, editors, {\em Proceedings of the Annual Meeting of the Association for Computational Linguistics (ACL)}, 2023.

\bibitem{imagenet15russakovsky}
Olga Russakovsky, Jia Deng, Hao Su, Jonathan Krause, Sanjeev Satheesh, Sean Ma, Zhiheng Huang, Andrej Karpathy, Aditya Khosla, Michael Bernstein, Alexander~C. Berg, and Li~Fei-Fei.
\newblock {ImageNet Large Scale Visual Recognition Challenge}.
\newblock {\em International Journal of Computer Vision (IJCV)}, 115(3):211--252, 2015.

\bibitem{shaw2018self}
Peter Shaw, Jakob Uszkoreit, and Ashish Vaswani.
\newblock Self-attention with relative position representations.
\newblock {\em arXiv preprint arXiv:1803.02155}, 2018.

\bibitem{su2024roformer}
Jianlin Su, Murtadha Ahmed, Yu~Lu, Shengfeng Pan, Wen Bo, and Yunfeng Liu.
\newblock Roformer: Enhanced transformer with rotary position embedding.
\newblock {\em Neurocomputing}, 568:127063, 2024.

\bibitem{DBLP:conf/icml/TouvronCDMSJ21}
Hugo Touvron, Matthieu Cord, Matthijs Douze, Francisco Massa, Alexandre Sablayrolles, and Herv{\'{e}} J{\'{e}}gou.
\newblock Training data-efficient image transformers {\&} distillation through attention.
\newblock In Marina Meila and Tong Zhang, editors, {\em International Conference on Machine Learning (ICML)}, 2021.

\bibitem{touvron2023llama}
Hugo Touvron, Louis Martin, Kevin Stone, Peter Albert, Amjad Almahairi, Yasmine Babaei, Nikolay Bashlykov, Soumya Batra, Prajjwal Bhargava, Shruti Bhosale, et~al.
\newblock Llama 2: Open foundation and fine-tuned chat models.
\newblock {\em arXiv preprint arXiv:2307.09288}, 2023.

\bibitem{van2017neural}
Aaron Van Den~Oord, Oriol Vinyals, et~al.
\newblock Neural discrete representation learning.
\newblock {\em Advances in Neural Information Processing Systems (NeurIPS)}, 2017.

\bibitem{vaswani2017attention}
Ashish Vaswani, Noam Shazeer, Niki Parmar, Jakob Uszkoreit, Llion Jones, Aidan~N. Gomez, Lukasz Kaiser, and Illia Polosukhin.
\newblock Attention is all you need.
\newblock In Isabelle Guyon, Ulrike von Luxburg, Samy Bengio, Hanna~M. Wallach, Rob Fergus, S.~V.~N. Vishwanathan, and Roman Garnett, editors, {\em Advances in Neural Information Processing Systems (NeurIPS)}, 2017.

\bibitem{wang2019encoding}
Benyou Wang, Donghao Zhao, Christina Lioma, Qiuchi Li, Peng Zhang, and Jakob~Grue Simonsen.
\newblock Encoding word order in complex embeddings.
\newblock In {\em International Conference on Learning Representations (ICLR)}, 2019.

\bibitem{wu2025scaling}
Chen Wu and Yin Song.
\newblock Scaling context, not parameters: Training a compact 7b language model for efficient long-context processing.
\newblock {\em arXiv preprint arXiv:2505.08651}, 2025.

\bibitem{yang2025qwen3}
An~Yang, Anfeng Li, Baosong Yang, Beichen Zhang, Binyuan Hui, Bo~Zheng, Bowen Yu, Chang Gao, Chengen Huang, Chenxu Lv, et~al.
\newblock Qwen3 technical report.
\newblock {\em arXiv preprint arXiv:2505.09388}, 2025.

\bibitem{yang2025RoPE}
Bowen Yang, Bharat Venkitesh, Dwarak Talupuru, Hangyu Lin, David Cairuz, Phil Blunsom, and Acyr Locatelli.
\newblock Rope to nope and back again: A new hybrid attention strategy.
\newblock {\em arXiv preprint arXiv:2501.18795}, 2025.

\bibitem{zhao2023length}
Liang Zhao, Xiaocheng Feng, Xiachong Feng, Bin Qin, and Ting Liu.
\newblock Length extrapolation of transformers: A survey from the perspective of position encoding.
\newblock {\em arXiv preprint arXiv:2312.17044}, 2023.

\bibitem{NEURIPS2024_724be447}
Lianmin Zheng, Liangsheng Yin, Zhiqiang Xie, Chuyue Sun, Jeff Huang, Cody~Hao Yu, Shiyi Cao, Christos Kozyrakis, Ion Stoica, Joseph~E. Gonzalez, Clark Barrett, and Ying Sheng.
\newblock Sglang: Efficient execution of structured language model programs.
\newblock In A.~Globerson, L.~Mackey, D.~Belgrave, A.~Fan, U.~Paquet, J.~Tomczak, and C.~Zhang, editors, {\em Advances in Neural Information Processing Systems}, volume~37, pages 62557--62583. Curran Associates, Inc., 2024.

\bibitem{zhou2024transfusion}
Chunting Zhou, Lili Yu, Arun Babu, Kushal Tirumala, Michihiro Yasunaga, Leonid Shamis, Jacob Kahn, Xuezhe Ma, Luke Zettlemoyer, and Omer Levy.
\newblock Transfusion: Predict the next token and diffuse images with one multi-modal model.
\newblock {\em arXiv preprint arXiv:2408.11039}, 2024.

\bibitem{zhu2023pose}
Dawei Zhu, Nan Yang, Liang Wang, Yifan Song, Wenhao Wu, Furu Wei, and Sujian Li.
\newblock Pose: Efficient context window extension of llms via positional skip-wise training.
\newblock {\em arXiv preprint arXiv:2309.10400}, 2023.

\end{thebibliography}
\bibliographystyle{plain}


\clearpage
\appendices

\section{Implementation Details of SeqPE}
\label{app:seqpe_impl}

\subsection{Example of 1D Position Representation\label{app:seqpe_embed_1d}}
In Figure \ref{fig:illustration}, we demonstrate how to represent 2-dimensional positions using our framework. Here, we further explain the use case for 1-dimensional positions.
For instance, when representing the position $123$ using our framework, we first convert it into a sequence $\boldsymbol{s} = (s_0^{0}=\text{1''}, s_1^{0}=\text{``2''}, \dots, s_k^{0}=\text{``3''})$, where the maximum number of digits $k$ is set to 3, and the data dimension $n$ is 1.
The embedding for the position tokens is given by $\mathbf{T} \in \mathbb{R}^{10 \times d}$, which is the same as in the 2-dimensional case. The embedding for positions of tokens in the sequence is represented by $\mathbf{O} \in \mathbb{R}^{3 \times d}$, corresponding to $k = 3$.
Finally, since $n = 1$ in this case, we have the embedding $\mathbf{D} \in \mathbb{R}^{1 \times d}$ for data dimensions. However, because $\mathbf{D}$ is the same for all positions in the 1-dimensional case, we simply fix the $\mathbf{D} = \mathbf{0}$ in practice.

\subsection{Regularization for Embedding Distances}

When implementing the contrastive-learning-based objective $\mathcal{L}_\delta$ in Equation~\ref{eq:contrastive}, the method used to construct the position set $\mathcal{C}_\delta$ with $m$ positions is critical. We implement two methods for constructing $\mathcal{C}_\delta$ and randomly select one of them for each training example. Suppose that we have a maximum position for training $\mathcal{L}_\delta$, denoted as $L_{max}$, which can be specified by the user. We first sample a pivot position $p \in [0, L_{max})$.

\myparagraph{Global Random Sampling.}
We randomly sample $k$ positions that are lexically similar to $p$. This is achieved by first converting $p$ to a string and then randomly applying one of the following operations: swapping two digits, removing a digit, or adding a digit. Next, we sample $m-k$ additional positions from $[0, L_{max})$ to construct the set $\mathcal{C}_\delta$, ensuring that $p^+ \in \mathcal{C}_\delta$ is selected from the $m-k$ sampled positions.

\myparagraph{Local Random Sampling.}
We observe that the global sampling method may make it difficult for the encoder to distinguish positions that are far apart. For example, the encoder should learn that the embedding of position 16384 should be closer to 123 than to 122. To address this, we propose an alternative method: we first randomly sample a small range $(p_{\text{left}}, p_{\text{left}} + \max(256, m))$, then sample all $m$ positions from this local range.

These sampling strategies can be applied in a similar manner to data of arbitrary dimensionality.

\subsection{Regularization for Out-of-Distribution Positions}

We implement a multi-head version of $\mathcal{L}_\delta$ in Equation~\ref{eq:distill}. Following the multi-head operation in Transformer model \cite{vaswani2017attention}, we split the embedding $\boldsymbol{e}_p \in \mathbb{R}^d$ of a position $p$ into $n$ heads:
$
\boldsymbol{e}_{p}^1, \dots, \boldsymbol{e}_{p}^n = \mathrm{SplitHeads}(\boldsymbol{e}_p),
$
where $\boldsymbol{e}_{p}^i\in \mathbb{R}^{d/n}$. We construct the $\boldsymbol{P}$ and $\boldsymbol{Q}$ similarity matrices for teacher and student positions individually for each head. The final loss of Equation~\ref{eq:distill} is averaged over all heads. Different from the objective $\mathcal{L}_\delta$ in Equation~\ref{eq:contrastive} that we only want the overall embedding are close in the representation space, we hope the embeddings of student positions could learn similar patterns as those of teachers at a fine-grained level. 

\subsection{Balancing Position Training Frequencies}
The training of $\mathcal{L}_{\delta}$ and $\mathcal{L}_{OOD}$ relies on data randomly sampled from $[0, L_{max})$. The choices of batch size for optimizing the two objectives will affect the training frequencies of each position given the fixed number of training steps. In our preliminary experiments, we find the batch sizes for $\mathcal{L}_{\delta}$ and $\mathcal{L}_{OOD}$ should be at least $\frac{16L_{max}}{10,000}$. In the text and image tasks, the training context length are $512$ and $14\times14$, and the hyper-parameter $L_{max}$ for the two objectives are set to $20,000$ and $100\times100$, respectively. For simplicity, we set the batch sizes of the two objectives to 32 for both
text and image tasks. 

\subsection{Random Shift\label{app:random_shift}}

In our work, we randomly shift the start positions of a small portion of the training data to stabilize the optimization of the position distillation objective $\mathcal{L}_\delta$ in Equation~\ref{eq:distill}. For example, we randomly change 10\% of the training positions from $(0, L)$ to $[z, L+z)$ for 1-dimensional data, where $L$ is the training context length and $z \in [0, L_{\text{max}} - L)$ is a randomly shifted start position.

Allowing this random shift is actually a feature of our \textsc{SeqPE}. Since most learnable position embeddings are implemented via lookup tables, training on shifted positions is not feasible. For strong baselines such as \textsc{ALiBi} and \textsc{RoPE}, they are in fact relative position encoding methods and are invariant to such shifts. Specifically, the core of \textsc{RoPE} is the Euler-formula-based rotary matrix $\mathbf{R}_{\Theta}$, under which $\mathbf{R}_{\Theta,i}^\top\mathbf{R}_{\Theta,j} = \mathbf{R}_{\Theta,j-i}$, as discussed in Section~\ref{sec:background}. Therefore, the position representations before and after shifting are identical for \textsc{RoPE}:
$$
\mathbf{R}_{\Theta,i}^\top\mathbf{R}_{\Theta,j} = \mathbf{R}_{\Theta,j-i} = \mathbf{R}_{\Theta,(j+z)-(i+z)} = \mathbf{R}_{\Theta,i+z}^\top\mathbf{R}_{\Theta,j+z}.
$$

Similarly, for \textsc{ALiBi}, according to Equation~\ref{eq:alibi_matrix}, we have:

$$
\mathbf{M}_{i, j} = -m(i-j) = -m(i+z-j-z) = \mathbf{M}_{i+z, j+z}.
$$

Therefore, shifting is a no-op for both \textsc{RoPE} and \textsc{ALiBi}. Here, we are not arguing that relative position encoding is inadequate, but rather aim for our model to learn everything directly from the data.

\begin{figure}[t]\centering
\tiny
\begin{minipage}{1.0\columnwidth}\vspace{0mm}    \centering
\begin{tcolorbox} 
    \raggedright
    \small
     \hspace{-6mm}
    \  \\
Answer the question based on the given documents. Only give me the answer and do not output any other words. \\
\ \\
The following are given documents. \\
\ \\
{\color{blue} \{MULTI\_DOCS\}} \\
\ \\
Answer the question based on the given documents. Only give me the answer and do not output any other words.\\
\ \\
Question: {\color{forestgreen} \{QUESTION\}} Answer:
{\color{red} \{ANSWER\}} \\
\end{tcolorbox}
    
\caption{The placeholder {\color{blue}MULTI\_DOCS} represents the long context with multiple Wikipedia documents, {\color{forestgreen}QUESTION} is the user question, and {\color{red}{ANSWER}} is the target answer. We evaluate performance by measuring the perplexity of the gold answer or generating a hypothesis using the fine-tuned language model at the position of {\color{red}{ANSWER}}. \label{fig:qa_prompt}}
\end{minipage}
 \vspace{-2em}
\end{figure}

\section{Implementation Details of Tasks}

\subsection{Language Modeling\label{app:task1_lm_impl}} 

The batch size per device is set to $32$ for training the language model \cite{radford2019language} on sequences with $512$ tokens. We train the model on 4 A6000 GPUs for 100K steps. The learning rate is $5 \times 10^{-5}$, with $4000$ warmup steps. We use a linear learning rate scheduler provided by \texttt{huggingface/transformers}. The dropout rate is set to $0.1$. We employ mixed-precision training using the BF16 format.

For evaluation, we compute perplexity on non-overlapping chunks with varied numbers of tokens on the validation and test sets, following the setup of \cite{press2021train}.

\subsection{Visualization for Effect of $\mathcal{L}_{\delta}$\label{app:visual_impl}}
We follow most of the setups in the language modeling task (\emph{i.e.}, Appendix~\ref{app:task1_lm_impl}) to visualize the effect of $\mathcal{L}_{\delta}$. When using $\mathcal{L}_{\delta}$, we set the $\alpha$ and $\beta$ in Equation~\ref{eq:final_obj} to $0.1$ and $0.0$, respectively. In contrast, both of them are set to $0.0$ when $\mathcal{L}_{\delta}$ is disabled.

\subsection{Long-context Question Answering\label{app:task2_qa_impl}} 
We retain most of the training settings from Appendix~\ref{app:task1_lm_impl}. However, for our QA task, we extend the training context length from $512$ to $1024$ to evaluate the adaptability of the PE methods. The pretrained model is fine-tuned for 10K steps, with $100$ warmup steps. We select the best model based on the perplexity of answer spans on the validation dataset.

\subsection{Image Classification\label{app:task3_imagenet}} 
We follow the standard training recipe from \cite{DBLP:conf/icml/TouvronCDMSJ21} to train the \textsc{ViT-S} model, except that we extend the training to $400$ epochs, as in \cite{heo2024rotary}. Training is conducted on machines equipped with 4 A100 or A6000 GPUs, with a batch size of $256$.

\end{document}